\documentclass[10pt,twocolumn,letterpaper]{article}

\usepackage{iccv}
\usepackage{times}
\usepackage{epsfig}
\usepackage{graphicx}
\usepackage{subcaption}
\usepackage{amsmath}
\usepackage{amssymb}
\usepackage{multirow}
\usepackage{booktabs}
\usepackage[dvipsnames]{xcolor}
\usepackage{colortbl}

\definecolor{tabhighlight}{HTML}{e5e5e5}
\newcommand{\tablestyle}[2]{\setlength{\tabcolsep}{#1}\renewcommand{\arraystretch}{#2}\centering\footnotesize}
\usepackage{pifont}
\newcommand{\cmark}{\ding{51}}%
\newcommand{\xmark}{\ding{55}}%

\usepackage[breaklinks=true,bookmarks=false]{hyperref}
\usepackage{graphicx}
\usepackage{pythonhighlight}

\iccvfinalcopy 

\def\iccvPaperID{****} 

\ificcvfinal\fi

\begin{document}

\title{COCO-O: A Benchmark for Object Detectors under Natural Distribution Shifts}

\author{%
  Xiaofeng Mao$^{\dagger}$ \quad Yuefeng Chen$^{\dagger}$ \quad Yao Zhu$^{\ddagger}$ \quad Da Chen$^{\S}$ \quad Hang Su$^{\mathparagraph}$\\ 
  \quad Rong Zhang $^{\dagger}$ \quad Hui Xue$^{\dagger}$\\
  $^{\dagger}$Alibaba Group, $^{\ddagger}$Zhejiang University, $^{\S}$University of Bath, $^{\mathparagraph}$Tsinghua University\\
  \texttt{\{mxf164419,yuefeng.chenyf\}@alibaba-inc.com} \\
}

\twocolumn[{%
\renewcommand\twocolumn[1][]{#1}%
\maketitle
\vspace{-12mm}
\begin{center}
    \centering
    \captionsetup{type=figure}
    \includegraphics[width=\textwidth]{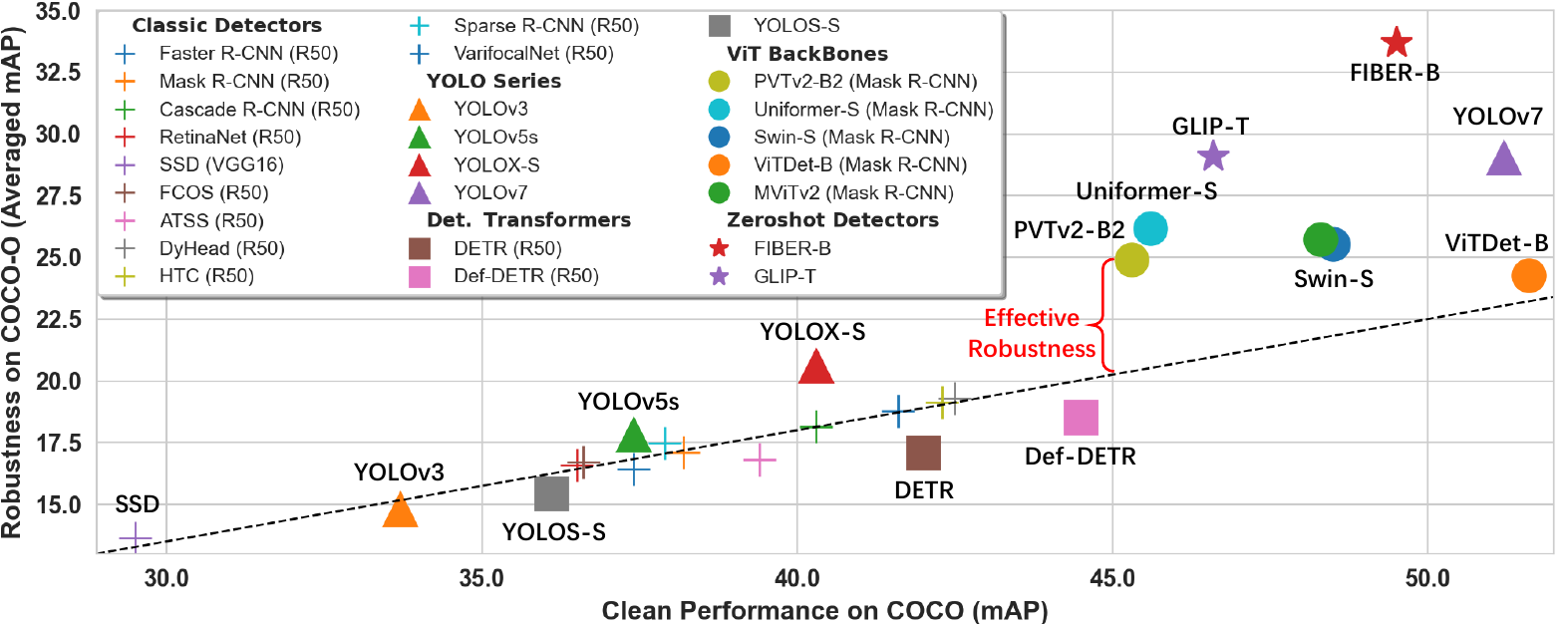}
    \vspace{-1.8em}
    \captionof{figure}{An overview of representative object detectors evaluated on COCO and our COCO-O dataset. The plotted dash line presents the linear function fitted on classic detectors, which demonstrates the trend of COCO-O mAP increasing along with COCO mAP. The \emph{Effective Robustness} (red text in figure) measures how far the model lies above the linear fit. }
    \label{fig:main}
\end{center}%
}]
\ificcvfinal\thispagestyle{empty}\fi

\begin{abstract}
Practical object detection application can lose its effectiveness on image inputs with natural distribution shifts. This problem leads the research community to pay more attention on the robustness of detectors under Out-Of-Distribution (OOD) inputs. Existing works construct datasets to benchmark the detector's OOD robustness for a specific application scenario, \eg, Autonomous Driving. However, these datasets lack universality and are hard to benchmark general detectors built on common tasks such as COCO.
To give a more comprehensive robustness assessment, we introduce \textbf{COCO-O}(ut-of-distribution), a test dataset based on COCO with 6 types of natural distribution shifts. COCO-O has a large distribution gap with training data and results in a significant \textbf{55.7\%} relative performance drop on a Faster R-CNN detector. We leverage COCO-O to conduct experiments on \textbf{more than 100 modern object detectors} to investigate if their improvements are credible or just over-fitting to the COCO test set. \textbf{Unfortunately, most classic detectors in early years do not exhibit strong OOD generalization.} We further study the robustness effect on recent breakthroughs of detector's architecture design, augmentation and pre-training techniques. Some empirical findings are revealed: 1) \textbf{Compared with detection head or neck, backbone is the most important part for robustness;} 2) \textbf{An end-to-end detection transformer design brings no enhancement, and may even reduce robustness;} 3) \textbf{Large-scale foundation models have made a great leap on robust object detection.} We hope our COCO-O could provide a rich testbed for robustness study of object detection. The dataset will be available at \url{https://github.com/alibaba/easyrobust/tree/main/benchmarks/coco_o}.

\end{abstract}

\let\thefootnote\relax\footnotetext{This research is supported in part by the National Key Research and Development Program of China under Grant No.2020AAA0140000.}

\section{Introduction}

Deep learning has achieved tremendous success in the field of computer vision. As a prerequisite, Deep Neural Networks (DNNs) rely on a rigorous assumption that training and testing data are independent and identically distributed. This ideal hypothesis is hardly satisfied in real-world applications, where the model may encounter data with distribution drift due to environmental changes, resulting in a significant decrease in performance and posing potential security issues. To solve this problem, the robustness study~\cite{hendrycks2021many,mao2022towards,taori2020measuring,chen2021amplitude} of DNNs under distribution shifts has emerged in the research area of image classification. 


However, most robustness researches merely focus on classification, and do not pay equal attention to other vision tasks, such as object detection. This phenomenon can be attributed to the lack of benchmark datasets. In contrast with holistic benchmarks~\cite{hendrycks2021natural,recht2019imagenet,wang2019learning,hendrycksbenchmarking,barbu2019objectnet} on ImageNet classification, the detection robustness benchmarks are limited. 
Previous work~\cite{michaelis2019benchmarking} benchmarks robustness using synthetic corruptions, however, it remains unclear if such simulated data can approximate real-world scenarios. Thus some other works collect images from internet to construct datasets.~\cite{chen2018domain,khodabandeh2019robust,he2019multi,lin2021domain,zhu2019adapting} use road scene datasets~\cite{yu2020bdd100k,cordts2016cityscapes,johnson2017driving,geiger2012we} to benchmark domain generalization of detectors. Such scene-specific dataset lacks universality and domain diversity, leading to a biased assessment of robustness. 
For evaluation on common tasks, ~\cite{zhao2022ood,inoue2018cross} collect natural OOD images based on PASCAL VOC~\cite{everingham2015pascal}. However, VOC is a small-scale detection task with limited number of categories, which has lagged behind the current standard evaluation protocol, \eg COCO~\cite{lin2014microsoft}, LVIS~\cite{gupta2019lvis} for detectors. 
We argue that more comprehensive and challenging benchmarks should be proposed to measure natural OOD robustness of modern detectors in 2020s.

In this work, we present COCO-O, a novel test dataset for COCO detection task which benchmarks robustness of object detectors under natural distribution shifts. 
COCO-O consists of 6,782 online-collected images belonging to 6 test domains: sketch, weather, cartoon, painting, tattoo and handmake. We compare our COCO-O with previous robust detection benchmarks in Table~\ref{tab:dataset-statistics}. 
Compared to VOC-related datasets, our COCO-O is more comprehensive with richer types of OOD shifts and larger dataset scale. COCO-O is fully compatible with the modern COCO evaluation protocol. Moreover, compared with COCO-related benchmarks, COCO-O is more challenging and can lead to 55.7\% relative performance drop on a Faster R-CNN detector. By calculating the Fréchet Inception Distance (FID)~\cite{heusel2017gans} to clean distribution, we show our COCO-O (with FID=132) has larger distribution shifts than COCO-C~\cite{michaelis2019benchmarking}.

\begin{table}[h!]
    \footnotesize
    \centering
    \tablestyle{5pt}{1.05}
  \begin{tabular}{l|c|c|c|c|c}
  \toprule
  \multirow{2}{*}{Datasets} & OOD & Class & Natural & Performance & \multirow{2}{*}{FID} \\
    & Types & Num. & Images & Drop (\%) & \\
  \midrule
  \rowcolor{tabhighlight} \multicolumn{6}{c}{VOC Scale Robustness Benchmarks} \\
    OOD-CV~\cite{zhao2022ood} & \textbf{5} & 10 & 2,632 & $\downarrow$ 26.6\% & 91 \\
   Clipart1k~\cite{inoue2018cross} & 1 & \textbf{20} & 1,000 & $\downarrow$ 59.8\% & \textbf{148} \\
   Watercolor2k~\cite{inoue2018cross} & 1 & 6 & 2,000 & $\downarrow$ 39.1\% & 113 \\
   Comic2k~\cite{inoue2018cross} & 1 & 6 & 2,000 & $\downarrow$ \textbf{71.5\%} & 147 \\
   \midrule
   \rowcolor{tabhighlight} \multicolumn{6}{c}{COCO Scale Robustness Benchmarks} \\
    COCO-C~\cite{michaelis2019benchmarking} & \textbf{15} & 80 & 0$^{*}$ & $\downarrow$ 49.8\% & 41 \\
   COCO-O (Ours) & 6 & 80 & \textbf{6,782} & $\downarrow$ \textbf{55.7\%} & \textbf{132} \\
   
  \bottomrule
  \end{tabular}
  \vspace{-3mm}
  \caption{Overview of existing general robust detection benchmark. $^{*}$Note that COCO-C has only synthetic images.}
  \label{tab:dataset-statistics}
  \vspace{-6mm}
\end{table}

Taking advantage from the proposed COCO-O, we additionally contribute extensive experiments on more than 100 modern object detectors to investigate the credibility of their reported improvements and whether they are just over-fitting to the COCO test set.
An overview of some key results is shown in Figure~\ref{fig:main}. Through a more precise \emph{Effective Robustness (ER)} metric~\cite{andreassenevolution} which eliminates extra impact brought by the variance of ID performance, we make a frustrating observation that most classic detectors have no great progress on robustness. However, recent breakthroughs in Visual Transformers (ViTs)~\cite{dosovitskiyimage} and large-scale vision foundation models have brought new hope for OOD robustness. Especially, zero-shot detectors~\cite{li2022grounded,doucoarse} pre-trained with massive image-language pairs exhibit great effectiveness on our COCO-O. Our results inspire future research to explore training data scaling or fusing external knowledge of human language to achieve more robust detection.
Besides, we analyse how OOD robustness is influenced by detector architecture, augmentation, pre-training, \textit{etc}. Some interesting findings are revealed, which can be summarized as: 1) Compared with the detection head or neck, backbone is the most important part for detector's robustness. Our empirical study shows scaling up backbone model or using advanced backbone design, \eg ResNeXt~\cite{xie2017aggregated}, Swin~\cite{liu2021swin} can bring greater robustness gains. 2) Detection transformers~\cite{carion2020end,zhudeformable} are more vulnerable than traditional non-end-to-end detectors under natural distribution shifts. Note that it is different from the previous experience~\cite{paul2022vision,bai2021transformers,naseer2021intriguing} in classification tasks, where ViTs are regarded as a robust learner. We hope our COCO-O could provide a rich testbed for robustness study of object detection, and we appeal that detection algorithms proposed in future should also evaluate their OOD generalization ability.

Our contributions are summarized below:
\begin{itemize}
    \item We propose COCO-O, the first COCO-scale test dataset for evaluating the robustness of detectors under natural distribution shifts.
    \item We benchmark the robustness of 100+ modern detectors and provide a thorough comparison in Section~\ref{exp:all}.
    \item Through analysing the impact factors of detector's robustness. We reveal some findings in Section~\ref{exp:analysis} that can help to develop more robust detection algorithms.

\end{itemize}

\begin{figure*}
  \centering
  \includegraphics[width=1.0\linewidth]{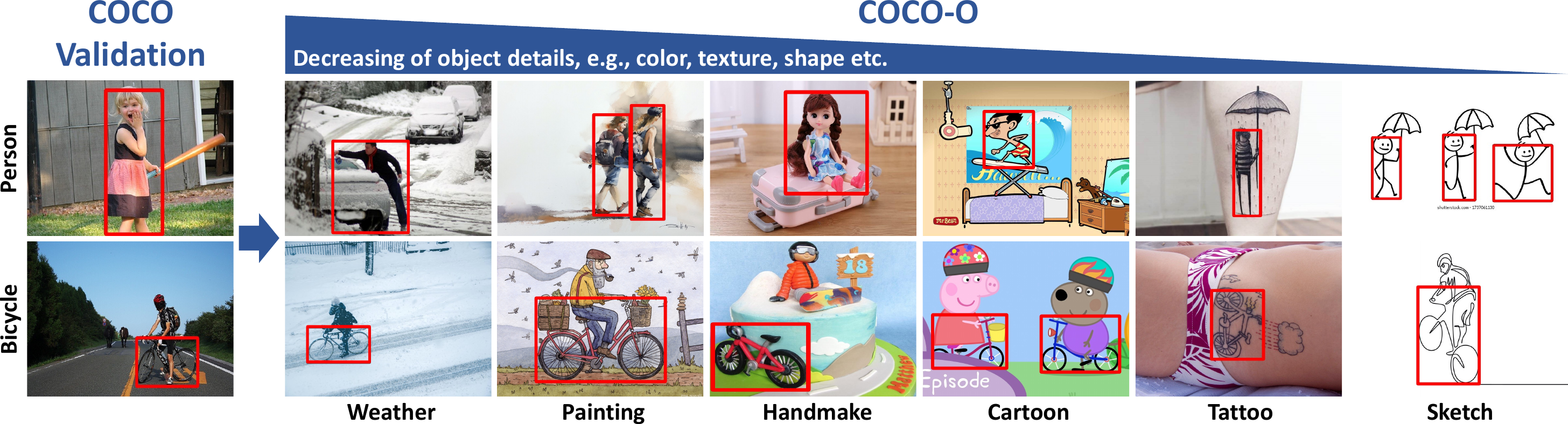}
  \vspace{-5mm}
  \caption{Visualization of our COCO-O. We adopt 6 domains, \textit{i.e.} weather, painting, handmake, cartoon, tattoo, sketch. The domains are ordered by decreasing details of their contained objects. Each domain presents an abstract levels of the objects.}
  \label{fig:samples}
  \vspace{-4mm}
\end{figure*}

\section{Related Work}
\paragraph{Object Detection}
Object detection task aims at classifying and localizing the objects in an image. Traditional detection methods can be divided into two categories: single-stage detectors~\cite{liu2016ssd,lin2017focal,zhou2019objects,tan2020efficientdet,tian2019fcos,law2018cornernet} and two-stage detectors~\cite{girshick2014rich,girshick2015fast,lin2017feature,ren2015faster,he2017mask,cai2019cascade,sun2021sparse}. There is also a research branch extended from single-stage detection, which utilizes lightweight design for real-time detection~\cite{redmon2016you,redmon2017yolo9000,redmon2018yolov3,ge2021yolox,bochkovskiy2020yolov4}. 
Recently, the success of transformer models~\cite{vaswani2017attention} in computer vision has led to the widespread use of transformer-based architectures~\cite{carion2020end,zhudeformable} in object detection, which replace the complex manual anchor design and non-maximum suppression procedure of previous methods.
In contrast to previous closed-set detection methods, open-set object detection~\cite{li2022grounded,wu2022grit} has also emerged as a mainstream research topic. By leveraging large-scale pre-training on image-language data, these methods can localize any object with only a given text description.

\paragraph{Robust Detection Benchmarks}
\label{related:benchmark}
Object detectors can fail under various conditions such as blur, occlusion, weather changes, deformation, \textit{etc}. To study the impact of these conditions, previous studies have constructed benchmark datasets via synthetic or online collected images. For instance, COCO-C~\cite{michaelis2019benchmarking} adds synthetic corruptions such as JPEG compression, gaussian noise to COCO~\cite{lin2014microsoft} test set. In this work, we do not consider image synthesis technique for benchmark construction since it has two inherent drawbacks: 1) it is hard to synthesize objects with pose or shape changes; 2) noise or artifacts will be introduced in synthesis process, leading to the deviation from natural image distribution. Another line of work proposes benchmarks for specific problems, \eg, environmental changes in autonomous driving~\cite{johnson2017driving}, object variation in aerial imagery~\cite{xia2018dota}, \textit{etc}. However we believe a general robustness benchmark should be built on some common detection tasks such as COCO or VOC~\cite{everingham2015pascal}.~\cite{zhao2022ood,inoue2018cross} collected OOD images based on VOC, but their task scale and domain diversity are still limited. To the best of our knowledge, COCO-O is the first natural OOD benchmark for COCO task. It has larger test set with more object categories and OOD types. 
\paragraph{Robust Detectors} Training robust detectors generalizing to unknown domain has been extensively studied in the literature. Most domain adaptation based methods~\cite{chen2018domain,inoue2018cross,he2019multi,khodabandeh2019robust,zhu2019adapting,kim2019diversify} require target domain data for adapting detectors. However, for online-deployed detectors, the test domain is open and indeterminate.~\cite{zhang2022towards} first studies the domain generalization problem in object detection. They eliminate the dependence within RoI features to improve the generalization of detection models under distribution shifts. To make detectors robust to image corruptions,~\cite{michaelis2019benchmarking} proposes to transfer styles of training images for data augmentation. Further, Det-AdvProp~\cite{chen2021robust} follows AdvProp~\cite{xie2020adversarial} to train detectors on clean and adversarial examples using two-way batchnorm. Such adversarially learned feature makes detector less sensitive to unknown distortions. Another branch of works~\cite{dong2022adversarially,zhang2019towards,chen2021class} aim at improving the adversarial robustness of detectors. Due to the well-known adversarial robustness and accuracy trade-off~\cite{zhang2019theoretically}, these methods suffer from a drop of clean mAP. Meanwhile, our experiment in Section~\ref{sec:4.2} suggests their OOD generalization ability has also decreased. 

\section{COCO-O}

\subsection{Choice of Test Domains}

As depicted in Figure~\ref{fig:samples}, in COCO-O, the selection of test domains is carried out by first dividing the objects into six abstract levels based on decreasing levels of details such as color, texture, and shape. For each abstract level, an appropriate domain is chosen. Most domain designs are motivated by ImageNet-R~\cite{hendrycks2021many}. We introduce them as follows: 1) Weather contains objects in challenging weather conditions, \eg rain, snow and fog. It is the easiest domain which has only appearance-based shifts and reserves most of the object details; 2) Painting includes most watercolor paintings which provide a realistic description of objects in a different image style; 3) Handmake consists of real-world human handicrafts, \eg, origami, toy, sculpture, \textit{etc}. The material of the object is changed in this domain; 4) Cartoon has images of 2D or 3D digital animation. It only preserves the rough structure and color information of the object. 5) Tattoo involves art drawing on human bodies. It can even include less image details than cartoon images, and some tattoos are black-only; 6) Sketch is considered the hardest case in COCO-O. It contains a set of line-drawing images missing texture and color. As a high-level abstraction of objects, detecting sketch objects requires more external knowledge or human priors. It should be noted that since traditional factors of small size objects, occlusion, illumination, image quality has been studied before~\cite{saleh2021occlusion,hu2017finding,xu2022rfla,chen2021exploring,haris2021task}, we do not adopt them as an individual test domain in COCO-O, but implicitly include them (Figure~\ref{fig:difficulty}). For instance, cars on a rainy night have poor illumination conditions or bicycles covered by snow are seriously occluded.


\subsection{Data Collection}
\label{sec:3.3}
We collect COCO-O images by searching the internet using a combinations of OOD scenario keywords and object categories from COCO. For instance, ``cartoon + dog" aims to gather a collection of animation dog images. Generally, most images searched by ``cartoon + dog'' are iconic~\cite{berg2009finding}, where single high quality object is centered in the image and can be localized easily. To obtain more non-iconic images, we follow the way used in COCO~\cite{lin2014microsoft} and add more object categories into keywords combinations, such as ``cartoon + dog + car''. We manually control the number of images retrieved by each keyword combination to ensure a balance among categories. For combinations that return only a few images, such as ``fog + bowl + tv'', we try to use multiple search engines for collecting more images. A list of the search queries is provided in Supplementary G.

\subsection{Dataset Statistics}
The annotated COCO-O has a total of 6,782 images and 26,624 labelled bounding boxes. It includes six test domains: Sketch (992 images, 3,707 objects), Weather (961 images, 4,509 objects), Cartoon (1,996 images, 8,774 objects), Painting (954 images, 4,879 objects), Tattoo (918 images, 1,489 objects) and Handmake (961 images, 3,266 objects). Original 80 COCO categories are adopted in our dataset. We additionally visualize the number of instances per image and class distribution in Supplementary A. Compared to COCO, COCO-O has roughly 5\% more images with only one single object, which may introduce potential gaps. However, the analysis in Supplementary A has demonstrated that the performance change brought by more iconic images can be negligible.

\begin{figure}
  \centering
  \includegraphics[width=0.96\linewidth]{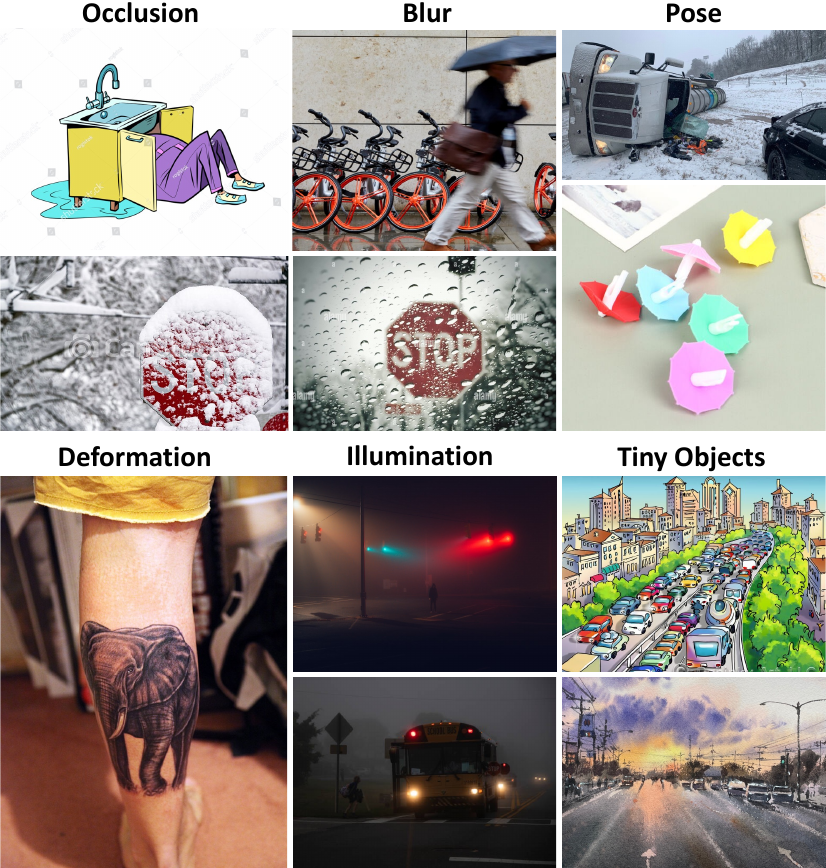}
  \vspace{-5pt}
  \caption{Some potential difficulties in COCO-O dataset.}
  \label{fig:difficulty}
  \vspace{-4mm}
\end{figure}

\subsection{Potential Difficulties in COCO-O}
\label{sec:3.4}
COCO-O is a challenging benchmark as it not only has large distribution shifts, but also contains some potential hard cases frequently encountered in detection tasks. We provide visualizations of these potential challenges in Figure~\ref{fig:difficulty}, most of which are specific to a particular test domain. For instance, weather changes bring additional difficulties for the detector, such as object occlusion caused by snow cover, poor illumination conditions in fog. In tattoo domain, geometric deformation caused by human bodies can pose an extra challenge. These potential challenges further enhance the difficulty of COCO-O. A robust detector should not only maintain consistent performance under distribution shifts, but also be able to tackle these potential challenges. 

\subsection{Other Applicable Tasks}
In addition to evaluating the OOD robustness of detectors, our dataset has potential applications for other detection-related tasks. Given the domain labels provided in COCO-O, one direct application is domain adaptation~\cite{oza2021unsupervised} or generalization~\cite{zhang2022towards} research. Few-shot learning~\cite{chen2021self} and incremental learning~\cite{wei2022incremental} are alternative solutions for OOD problem. Specifically, COCO-O can also be leveraged for cross-domain few-shot detection~\cite{gao2022acrofod,guo2020broader}, where only a limited number of samples are available for training detectors. Overall, our dataset offers abundant resources and challenges, and facilitates the advancement of various tasks in the object detection field.

\begin{table*}\centering
\scriptsize
\begin{tabular}{lcc|c|ccccccc|c}
\toprule
& & & COCO & \multicolumn{7}{c|}{COCO-O (mAP)} & Effective \\
& & & mAP & Sketch & Weather & Cartoon & Painting & Tattoo & Handmake & Avg. & Robustness\\
\midrule
\multirow{13}{*}{\rotatebox{90}{Faster R-CNN}} & \multirow{5}{*}{BackBone} & RN-50~\cite{he2016deep} & 37.4 & 9.8 & 25.1 & 13.9 & 23.3 & 10.1 & 16.3 & 16.4 & \textcolor{red}{-0.41} \\
& & RN-101~\cite{he2016deep} & 39.4 & 10.7	& 30.3 & 16.1 & 27.6 & 11.3 & 18.6 & 19.1 & \textcolor{green!50!black}{+1.37} \\
& & RX-101-32x4d~\cite{xie2017aggregated} & 41.2 & 12.1	& 31.6 & 16.3 & 28.9 
 & 11.3 &	19.7 &	20.0 & \textcolor{green!50!black}{+1.44}\\
& & Swin-T~\cite{liu2021swin} & 42.5 & 12.8 &	34.2 &	17.3 &	31.1 &	10.6 &	20.4 &	21.1 & \textcolor{green!50!black}{+1.94} \\
& & PVTv2-B2~\cite{wang2022pvt} & 45.6 & 16.3 &	37.8 &	22.1 &	35.9 &	13.0 &	24.0 &	24.9 & \textcolor{green!50!black}{+4.33} \\
\cmidrule{2-12}
& \multirow{3}{*}{Neck} & FPN~\cite{lin2017feature} & 37.4 & 9.8 & 25.1 & 13.9 & 23.3 & 10.1 & 16.3 & 16.4 & \textcolor{red}{-0.41} \\
& & PAFPN~\cite{liu2018path} & 37.5 & 9.4 &	26.9 &	13.9 &	23.8 &	10.7 &	15.8 &	16.8 & \textcolor{red}{-0.13} \\
& & NAS-FPN~\cite{ghiasi2019fpn} & 38.0 & 8.7 &	23.8 &	12.7 &	22.1 &	8.7 &	14.6 &	15.1 & \textcolor{red}{-2.00} \\
\cmidrule{2-12}
& \multirow{5}{*}{Det. Head} & Standard & 37.4 & 9.8 & 25.1 & 13.9 & 23.3 & 10.1 & 16.3 & 16.4 & \textcolor{red}{-0.41} \\
& & Cascade~\cite{cai2019cascade} & 40.3 & 10.8 &	27.8 &	14.9 &	25.5 &	12.5 &	17.4 &	18.2 & \textcolor{green!50!black}{+0.02} \\
& & SABL~\cite{wang2020side} & 39.9 & 10.6 &	27.3 &	15.0 &	25.3 &	11.8 &	18.2 &	18.0 & \textcolor{green!50!black}{+0.08}   \\
& & 2\_Heads~\cite{wu2020rethinking} & 40.0 & 10.6 &	30.4 &	14.6 &	25.7 &	12.0 &	18.4 &	18.6 & \textcolor{green!50!black}{+0.62}  \\
& & Groie~\cite{rossi2021novel} & 38.3 & 9.8 &	28.1 &	14.0 &	24.4 &	10.8 &	16.2 &	17.2 & \textcolor{red}{-0.02}   \\

\midrule

\multirow{12}{*}{\rotatebox{90}{RetinaNet}} & \multirow{5}{*}{BackBone} & RN-50~\cite{he2016deep} & 36.5 & 9.8 &	25.9 &	13.8 &	23.7 &	10.4 &	16.0 &	16.6 & \textcolor{green!50!black}{+0.18} \\
& & RN-101~\cite{he2016deep} & 38.5 & 10.9 &	30.2 &	15.3 &	27.3 &	11.6 &	18.9 &	19.0 & \textcolor{green!50!black}{+1.71} \\
& & RX-101-32x4d~\cite{xie2017aggregated} & 39.9 & 12.2 &	32.1 &	16.1 &	28.0 &	11.2 &	19.7 &	19.9 & \textcolor{green!50!black}{+1.93} \\
& & Swin-T~\cite{liu2021swin} & 41.4 & 11.1 & 33.7 & 16.4 & 31.0 & 11.1 & 20.0 & 20.6 & \textcolor{green!50!black}{+1.92} \\
& & PVTv2-B2~\cite{wang2022pvt} & 44.6 & 17.6 & 38.9 & 22.0 & 35.1 & 14.2 & 23.2 & 25.2 & \textcolor{green!50!black}{+5.10} \\
\cmidrule{2-12}
& \multirow{3}{*}{Neck} & FPN~\cite{lin2017feature} & 36.5 & 9.8 &	25.9 &	13.8 &	23.7 &	10.4 &	16.0 &	16.6 & \textcolor{green!50!black}{+0.18} \\
& & PAFPN~\cite{liu2018path} & 36.7 & 9.5 & 27.0 & 13.5 & 24.5 & 10.9 & 16.2 & 16.9 & \textcolor{green!50!black}{+0.42} \\
& & NAS-FPN~\cite{ghiasi2019fpn} & 36.1 & 9.0 & 27.3 & 11.5 & 21.7 & 8.7 & 13.9 & 15.4 & \textcolor{red}{-0.90} \\
\cmidrule{2-12}
& \multirow{4}{*}{Det. Head} & Standard & 36.5 & 9.8 &	25.9 &	13.8 &	23.7 &	10.4 &	16.0 &	16.6 & \textcolor{green!50!black}{+0.18} \\
& & SABL~\cite{wang2020side} & 37.7 & 9.0 & 26.3 & 13.3 & 24.1 & 11.6 & 16.4 & 16.8 & \textcolor{red}{-0.18} \\
& & FSAF~\cite{zhu2019feature} & 37.4 & 9.4 & 25.2 & 13.5 & 23.2 & 11.9 & 15.9 & 16.5 & \textcolor{red}{-0.31} \\
& & FreeAnchor~\cite{zhang2019freeanchor} & 38.7 & 10.2 & 26.3 & 14.3 & 24.6 & 12.1 & 16.5 & 17.3 & \textcolor{red}{-0.08}  \\
\midrule
\multicolumn{3}{c|}{FCOS~\cite{tian2019fcos}} & 36.6 & 9.8 & 25.9 & 13.3 & 24.2 & 10.4 & 16.7 & 16.7 & \textcolor{green!50!black}{+0.25}  \\
\midrule
\multicolumn{3}{c|}{DETR~\cite{carion2020end}} & 42.0 & 9.0 & 30.0 & 12.3 & 23.9 & 11.6 & 15.7 & 17.1 & \textcolor{red}{-1.82}  \\
\multicolumn{3}{c|}{Deformable DETR~\cite{zhudeformable}} & 44.5 & 10.5 & 30.2 & 15.1 & 26.2 & 10.6 & 18.6 & 18.5 & \textcolor{red}{-1.49}  \\

\bottomrule
 
\end{tabular}
\caption{Comparison of object detectors with different backbone, neck and head. The effective robustness lower than the linear trend in Figure~\ref{fig:main} is highlighted by red. Number in green means the model is above the linear trend. }
\vspace{-10pt}
\label{tab:arch}
\end{table*}

\section{Experiments}
\label{exp:all}
In Section~\ref{exp:analysis}, we study how some basic components, such as detection architecture, augmentation and pre-training effect on the OOD robustness of traditional detection algorithms. We utilize COCO-O to examine previously proposed robust and SOTA detectors in Section~\ref{sec:4.2} and~\ref{sec:4.3}.

\paragraph{Experimental Setup.} 
To study the robustness effect of detection architecture and pre-training, we adopt typical two-stage Faster R-CNN~\cite{ren2015faster} and one-stage RetinaNet~\cite{lin2017focal} for baselines. For analyzing the impact of data augmentation on robustness, YOLOX~\cite{ge2021yolox} is used as baseline. All the above experiments are implemented by mmdetection~\cite{chen2019mmdetection} and use the consistent training configuration for fair comparison. Besides, for benchmark experiments on robust and SOTA detectors, we by default adopt the optimal training settings reported in their papers. Pre-trained models are directly used for robust and SOTA detection methods who have released their official weights.

\paragraph{Metrics.}\label{sec:exp:metric} A frequently-used metric of object detection is the mean Average Precision (mAP) which averages over Intersection over Unions (IoUs) between 50\% and 95\%. Similarly, we also adopt mAP to measure the robustness on each OOD case in COCO-O. The averaged mAP on 6 test domains is used as the overall performance. To exclude the impact of the linear trend of performance improvement on in- versus out-of-distribution data, we also adopt the Effective Robustness (ER) metric proposed by~\cite{andreassenevolution}. Given a set of classic detectors $\mathcal{F}$, we approximate the linear trend by $\beta(\textbf{mAP}_{id}(\cdot))$, where $\textbf{mAP}_{id}(\cdot)$ is the COCO mAP metric, $\beta$ is a learnable linear function fitted using observations on $\mathcal{F}$. Thus for any detector $f$, the performance on COCO-O can be predicted by $\beta(\textbf{mAP}_{id}(f))$. We use Scipy~\cite{virtanen2020scipy} for linear regression on 11 classic detectors in Figure~\ref{fig:main}. The slope is calculated as 0.45. Finally the effective robustness of a detector $f$ can be defined as:
\begin{equation}
    \textbf{ER}(f) = \textbf{mAP}_{ood}(f)-0.45 \times \textbf{mAP}_{id}(f),
\end{equation}
where $\textbf{mAP}_{ood}(\cdot)$ is our COCO-O mAP metric.

\subsection{Analysis of Traditional Detectors}
\label{exp:analysis}

\paragraph{Robustness vs. Detection Architecture.}Classic object detectors consist of three components: backbone, neck and head, each of which has a distinct role in feature extraction, feature map fusion, object localization and classification. In order to investigate the impact of each component on the overall robustness, we adopt default setting with ResNet-50~\cite{he2016deep} backbone, FPN~\cite{lin2017feature} neck, standard head for baseline and modify each part using some advanced designs. For backbone, we compare three different architectures: ResNet series~\cite{he2016deep}, ViT-based Swin-T~\cite{liu2021swin} and PVTv2-B2~\cite{wang2022pvt}. For neck, FPN~\cite{lin2017feature} and its two variants: PAFPN~\cite{liu2018path}, NAS-FPN~\cite{ghiasi2019fpn} are compared. For detection head, we analyse 3 and 4 advanced head designs for single- and two-stage detectors respectively. The results are reported in Table~\ref{tab:arch}. Surprisingly, it suggests that advanced detection architectures with higher clean COCO mAP do not imply better robustness. Some methods, such as NAS-FPN~\cite{ghiasi2019fpn}, FSAF~\cite{zhu2019feature} even 
make the model more fragile under natural distribution shifts. The best architectures of neck and head on Faster R-CNN are PAFPN~\cite{liu2018path} and 2\_Heads~\cite{wu2020rethinking}, which yield -0.13 and +0.62 effective robustness. Such improvement is marginal, indicating that advanced techniques on neck and head have a limited effect on robustness. In contrast, the backbone plays a more important role. Simply replacing ResNet50~\cite{he2016deep} with PVTv2-B2~\cite{wang2022pvt} can achieve +8.5 and +8.6 COCO-O mAP, +4.33 and +5.10 effective robustness on Faster R-CNN and RetinaNet respectively. Motivated by this phenomenon, for the first time we emphasize the vital role of feature extractor in detectors to enhance OOD robustness. 

In addition to classic detectors, some works have innovated the entire detection framework by modeling localization with point prediction~\cite{law2018cornernet} or self-attention~\cite{carion2020end}. It is still lacking sufficient robustness analysis on these methods. We evaluate the OOD robustness of three novel frameworks, namely FCOS~\cite{tian2019fcos}, DETR~\cite{carion2020end}, Deformable DETR~\cite{zhudeformable} on COCO-O. However, the results of Table~\ref{tab:arch} show that most frameworks do not bring a great promotion of effective robustness. An interesting phenomenon is that detection transformers have the worst OOD generalization ability, contrary to the general conclusion in the field of image classification that transformers can enhance OOD robustness~\cite{paul2022vision,bai2021transformers,naseer2021intriguing}.

\begin{figure}
  \centering
  \includegraphics[width=0.93\linewidth]{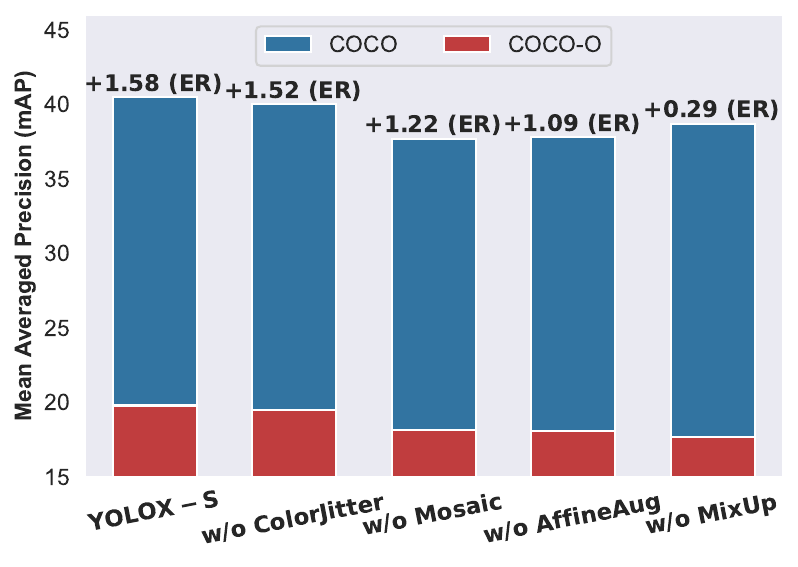}
  \vspace{-3mm}
  \caption{The robustness effect of different used augmentations in YOLOX-S detector.}
  \label{fig:augment}
  \vspace{-4mm}
\end{figure}

\paragraph{Robustness vs. Augmentations.}Data augmentation is a widely used technique for enhancing generalization. Especially in object detection, where heavy augmentations have been a crucial factor for the success of YOLO series models. To study its effect on OOD robustness, we adopt YOLOX-S~\cite{ge2021yolox}, one of the strongest detectors which adopts diverse augmentations including MixUp~\cite{zhangmixup}, ColorJitter, Mosaic~\cite{bochkovskiy2020yolov4}, Random Affine, \textit{etc}. Then we re-train the detector by iteratively deleting one augmentation to observe the variance of robustness. In Figure~\ref{fig:augment}, it can be suggested that all used augmentations contribute to the enhanced robustness. Among them, MixUp plays a principal role for OOD robustness. Removing it causes a significant drop in mAP on COCO-O from 19.8 to 17.7, as well as a decrease in effective robustness. This implies that, in addition to classification tasks, MixUp can also help for domain generalization on object detection. By comparison, ColorJitter is the least effective augmentation, it merely promotes 0.06 ER and 0.3 mAP on COCO-O.

\begin{table*}\centering
\scriptsize
\begin{tabular}{lcc|c|c|ccccccc|c}
\toprule
&\multicolumn{2}{c|}{BackBone Pre-training} & Detector & COCO & \multicolumn{7}{c|}{COCO-O (mAP)} & Effective \\
& Method & Data & Pre-training & mAP & Sketch & Weather & Cartoon & Painting & Tattoo & Handmake & Avg. & Robustness\\
\midrule
\multirow{7}{*}{\rotatebox{90}{Faster R-CNN}} & - & - & - & 37.2 & 10.6 & 23.3 & 14.6 & 23.1 & 9.2 & 16.0 & 16.1 & \textcolor{red}{-0.65} \\
& Sup & IN-1K & - & 37.4 & 9.8 & 25.1 & 13.9 & 23.3 & 10.1 & 16.3 & 16.4 & \textcolor{red}{-0.41} \\
& Sup\_RSB~\cite{wightman2021resnet} & IN-1K & - & 40.8 & 11.0	& 29.8 & 14.5 & 27.3 & 12.2 & 18.9 & 19.0 & \textcolor{green!50!black}{+0.59} \\
& SwAV~\cite{caron2020unsupervised} & IN-1K & - & 38.6 & 8.3 & 26.5 & 11.4 & 22.3 & 8.8 & 15.7 & 15.5 & \textcolor{red}{-1.9} \\
& MoCov2~\cite{chen2020improved} & IN-1K & - & 37.5 & 9.8	& 26.8 & 13.8 & 22.8 & 10.1 & 16.6 & 16.7 & \textcolor{red}{-0.23} \\
& Sup & IN-21K~\cite{ridnik1imagenet} & - & 38.9 & 9.9	& 30.1 & 14.8 & 26.0 & 11.8 & 19.6 & 18.7 & \textcolor{green!50!black}{+1.20} \\
& Sup & IN-1K & Obj365~\cite{shao2019objects365} & 42.1 & 11.5	& 30.1 & 16.4 & 27.0 & 11.3 & 19.3 & 19.3 & \textcolor{green!50!black}{+0.32} \\

\midrule
\multirow{6}{*}{\rotatebox{90}{RetinaNet}} & Sup & IN-1K & - & 36.5 & 9.8 & 25.9 & 13.8 & 23.7 & 10.4 & 16.0 & 16.6 & \textcolor{green!50!black}{+0.18} \\
& Sup\_RSB~\cite{wightman2021resnet} & IN-1K & - & 39.0 & 11.0	& 28.4 & 14.5 & 25.3 & 11.4 & 17.7 & 18.1 & \textcolor{green!50!black}{+0.50} \\
& SwAV~\cite{caron2020unsupervised} & IN-1K & - & 38.7 & 8.5	& 28.4 & 11.3 & 22.2 & 9.4 & 16.0 & 16.0 & \textcolor{red}{-1.45} \\
& MoCov2~\cite{chen2020improved} & IN-1K & - & 36.2 & 11.2	& 26.6 & 13.1 & 23.6 & 10.0 & 15.1 & 16.6 & \textcolor{green!50!black}{+0.31} \\
& Sup & IN-21K~\cite{ridnik1imagenet} & - & 38.2 & 9.8	& 28.9 & 14.3 & 24.8 & 11.6 & 18.9 & 18.1 & \textcolor{green!50!black}{+0.86} \\
& Sup & IN-1K & Obj365~\cite{shao2019objects365} & 41.0 & 11.9 & 28.7 & 16.3 & 25.5 & 10.5 & 19.1 & 18.7 & \textcolor{green!50!black}{+0.22} \\

\bottomrule
 
\end{tabular}
\vspace{-1mm}
\caption{The reported COCO-O performance of detectors with different pre-training methods.}
\label{tab:pretrain}
\vspace{-4mm}
\end{table*}

\paragraph{Robustness vs. Pre-training.}``Pre-training and fine-tuning'' is still the de facto paradigm for object detection task. Although previous study~\cite{hendrycks2019using} have demonstrated the efficacy of pre-training for constructing reliable models, they have not considered object detection tasks. 
To build a detector, there will be multiple ways that 1) training from scratch without any pre-training data; 2) using pre-trained checkpoint to initialize the backbone; 3) leveraging datasets like Object365~\cite{shao2019objects365} to pre-train the overall detector and then fine-tuning. We list and compare the most pre-training settings in Table~\ref{tab:pretrain}. For detectors trained from scratch, we employ a longer 6$\times$ schedules adviced by~\cite{he2019rethinking}. Besides, all other compared methods are using ResNet-50 backbone and same training hyper-parameters for ensuring fairness. 
The results suggest that the detector can becomes more robust when ImageNet-1K is used for pre-training.  However the obtained robustness depends on an appropriate pre-training method. For example, we compare two self-supervised approaches for backbone pre-training: SwAV~\cite{caron2020unsupervised} and MoCov2~\cite{chen2020improved}, and find that SwAV pre-trained detector performs poorly on OOD robustness. Instead, an elaborate supervised pre-training procedure~\cite{wightman2021resnet} can even beat SwAV and MoCov2. Moreover, using more data for pre-training can further improve the robustness. It is foreseeable, but we need to remark that big data policy works in both backbone and detector training stage. On Fatser R-CNN, using backbone pre-trained on ImageNet-21K~\cite{ridnik1imagenet} and detector pre-trained on Object365~\cite{shao2019objects365} can get +1.20 and +0.32 effective robustness respectively. 

\begin{figure}
  \centering
  \includegraphics[width=1.0\linewidth]{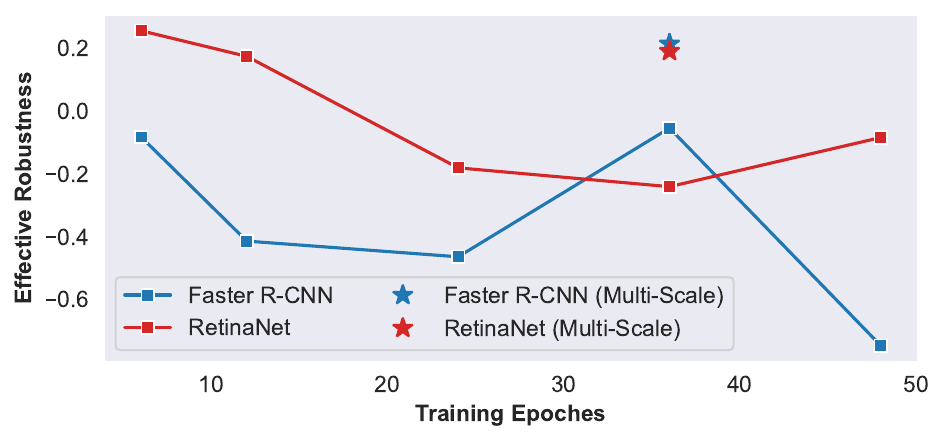}
  \vspace{-22pt}
  \caption{Multi-scale and longer training vs. robustness}
  \label{fig:epochs}
  \vspace{-4mm}
\end{figure}

\paragraph{Others. } In addition to the techniques mentioned above, there are many others tricks for improving detectors during training. In this study, we investigate the impact of two common practices, multi-scale training and longer-epoch training, on the robustness of detectors. The results are reported in Figure~\ref{fig:epochs}. Interestingly, we do not find any trend of robustness increasing with longer training times. It suggests the improvement of clean mAP brought by longer epochs of training may indicate over-fitting on the COCO test set. On the other hand, training with auxiliary multi-scale inputs can slightly improve the effective robustness of Faster R-CNN and RetinaNet by +0.26 and +0.43 respectively.

\subsection{Results on Robust Detectors}
\label{sec:4.2}
There have been several works proposed to train robust object detectors. For resisting adversarial attacks, RobustDet~\cite{dong2022adversarially} trains an adversarially robust object detector on clean and adversarial images via adversarially-aware convolution. Det-AdvProp~\cite{chen2021robust} also uses adversarial images for training but it aims at better clean performance through a separate batch norm design similar to AdvProp.~\cite{michaelis2019benchmarking} has shown that style transfer augmentation can significantly improve corruption robustness. However, most of these methods do not evaluate their models under natural distribution shifts due to the lack of OOD benchmark datasets on COCO. Take advantage from our COCO-O, in this work we present an assessment of their natural OOD robustness in Table~\ref{tab:robust_det}. Both Stylized-Aug and Det-AdvProp can improve performance on COCO-O effectively, but it should be note that the former method sacrifices 0.7 clean mAP while the latter even achieves +0.6. As an adversarial defense method, RobustDet has a greater impact on clean performance, meanwhile, such adversarially robust detectors have low generalization ability on our COCO-O.

\begin{table}[h!]
    \small
    \centering
    \tablestyle{5pt}{1.05}
  \begin{tabular}{c|c|c|c|c}
  \toprule
  Baseline & Robust & COCO & COCO-O & Effective \\
  Models & Methods & mAP & mAP & Robustness \\
  \midrule
   \multirow{2}{*}{Faster R-CNN} & - & 37.4 & 16.4 & \textcolor{red}{-0.41} \\
   & Stylized-Aug & 36.1 & 20.4 & \textcolor{green!50!black}{+4.12} \\
   \midrule
   \multirow{2}{*}{EfficientDet-D1} & - & 40.2 & 22.0 & \textcolor{green!50!black}{+3.86} \\
   & Det-AdvProp & 40.8 & 22.9 & \textcolor{green!50!black}{+4.56} \\
    \midrule
   \multirow{2}{*}{SSD} & - & 42.0 & 18.7 & \textcolor{red}{-0.18} \\
   & RobustDet & 31.0 & 13.7 & \textcolor{red}{-0.28} \\
  \bottomrule
  \end{tabular}
  \caption{The performance of robust detectors on COCO-O.}
  \vspace{-5mm}
  \label{tab:robust_det}
\end{table}

\begin{figure*}
  \centering
  \includegraphics[width=1.0\linewidth]{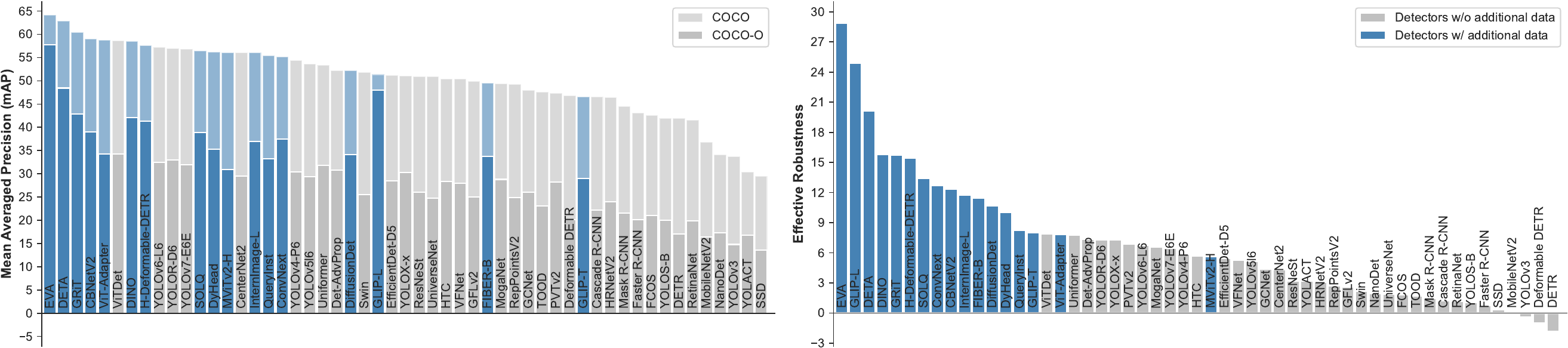}
  \vspace{-6mm}
  \caption{The reported COCO-O mAP (left) and effective robustness (right) on 53 SOTA detectors.}
  \label{fig:sota}
  \vspace{-13pt}
\end{figure*}

\subsection{Results on SOTA Detectors} 
\label{sec:4.3}
To investigate whether the latest developments in the field of object detection have made progress in closing the OOD distortion robustness gap, we collected 53 powerful detectors based on the COCO Leaderboard\footnote{https://paperswithcode.com/sota/object-detection-on-coco} and evaluated their performance on our COCO-O. The core results are shown in Figure~\ref{fig:sota}. Currently, EVA~\cite{fang2022eva} stands out from thousands of open-sourced detectors and holds the first place on COCO-O. EVA is a billion-scale vision foundation model, which shows model\&data scaling is still the most direct and effective way towards OOD generalization in object detection. However, without the help of additional data, all detectors are facing a giant decline on COCO-O (shown in grey bars in Figure~\ref{fig:sota}). A frustrate situation is that the most effectively robust detector ViTDet-H~\cite{li2022exploring}, which is trained on standard data (COCO, ImageNet-1K), achieves merely 7.885 and even cannot enter the top@10 of the ranking list. This finding suggests that most recent progression may be due to the use of more training data. 

\section{Discussion}
\label{sec:discuss}

\paragraph{Large-scale foundation models have made the greatest progress in robust object detection. }Since CLIP~\cite{radford2021learning} has been firstly verified its success on comprehensive vision benchmarks, large-scale foundation models gradually become the mainstream of visual research.~\cite{geirhos2021partial} have shown classification models trained on billion-scale data achieve significant robustness under distribution shifts, and non-trivial progress on closing the gap between human and machine vision. Our paper discovers a similar phenomenon on object detection. Figure~\ref{fig:sota} shows large-scale pre-trained detectors have made considerable progress in OOD robustness. However, we must remain cautious as this success may be attributed to the fact that the detector has seen OOD data during training. For the research community, it is still more meaningful to focus on innovating robust detection algorithms rather than solely relying on larger training datasets.

\paragraph{Why DETRs underperform traditional detectors on robustness? }In our results, the DETRs~\cite{carion2020end} are shown to be vulnerable.
To achieve end-to-end detection, DETR introduces two major modifications: 1) replacing the conventional one-to-many label assignment with one-to-one Hungarian matching, and 2) learning a set of object queries for localization without any prior anchor information. We suspect that the label assignment rules or localization queries learned by DETR may heavily rely on the training data, which is unfavorable for generalization. Instead, human-designed priors could have stronger generalization and interpretability. Till now, the exact reason for DETR's poor robustness remains unsolved. Further research is needed to study this problem.

\begin{figure}[h!]
  \centering
  \includegraphics[width=0.95\linewidth]{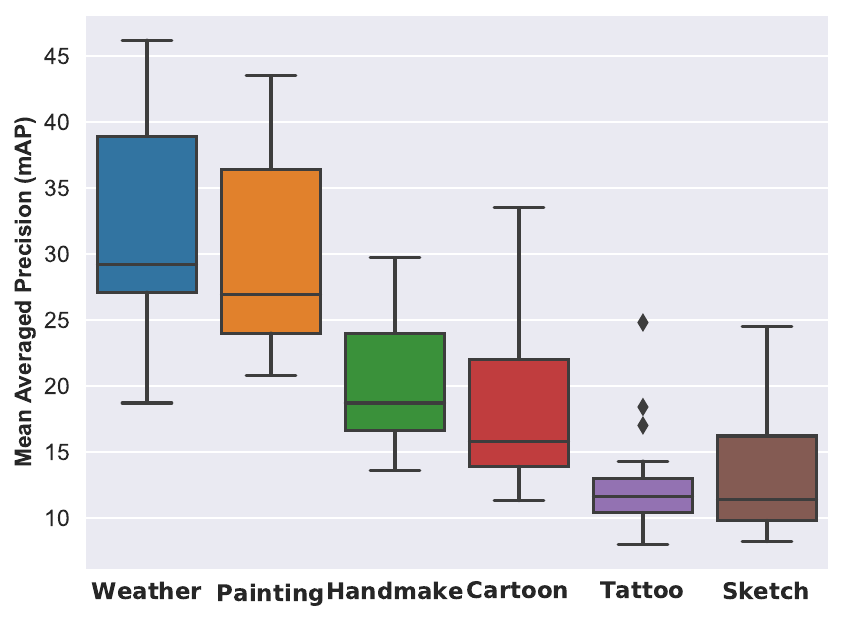}
  \vspace{-2.5mm}
  \caption{Robustness level on six test domains of COCO-O. The statistical result is counted from detectors in Figure~\ref{fig:main}. }
  \label{fig:domains}
  \vspace{-15pt}
\end{figure}

\paragraph{Analysis on different test domains.} By using COCO-O, we can study the influence of each type of natural distribution shift. We adopt detectors in Figure~\ref{fig:main}, and evaluate them on each test domain of COCO-O. The results are shown in Figure~\ref{fig:domains}. Same with the order in Figure~\ref{fig:samples}, sketch and tattoo objects are the hardest to detect, as they lost important feature \eg colors for detection. In contrast, appearance-based shifts such as weather are relatively easier to handle. Detectors have the lowest performance variance on tattoo objects, most of them have the mAP below 13. 
It suggests the necessity of designing specialized detectors for tattoo images.

\paragraph{Our difference with COCO-C. }People may concern about the necessity of our COCO-O since previous COCO-C~\cite{michaelis2019benchmarking} has been taken as a generally accepted robustness metric for object detection. Here we must reaffirm our difference and superiority with COCO-C: 1) COCO-C is a synthetic dataset and has limitations as discussed in Section~\ref{related:benchmark}. In contrast, COCO-O includes realistic images, which are more representative of the real-world scenarios; 2) The purposes are different. COCO-C measures robustness under image corruptions, while corruptions may not cover all real-world OOD shifts such as artificial creation in different styles and forms. COCO-O contains more diverse realistic images that cover such OOD shifts; 3) As demonstrated in Supplementary C, our COCO-O has a lower correlation with COCO mAP. It implies that our proposed COCO-O, which covers robustness evaluation scenarios that are not considered in COCO validation set, can be a meaningful metric complement with COCO mAP to reflect the overall performance. 

\section{Conclusion}
In this paper, we propose a novel dataset called COCO-O to benchmark object detection under natural distribution shifts. With a thorough diagnosis of more than 100 modern object detectors, we demonstrate that detecting objects with OOD shifts remains a challenge and requires further attention from the research community. Additionally, we empirically investigate how OOD robustness is influenced by various factors, including detector architecture, augmentation, pre-training, \textit{etc}. With our COCO-O dataset, innovative techniques can be developed to enhance the OOD robustness of existing detection algorithms, which will be the focus of our future work.

{\small
\bibliographystyle{ieee_fullname}
\bibliography{egbib}
}

\clearpage
\appendix
\section{More Visualization of COCO-O}
\begin{figure}[h!]
  \centering
  \includegraphics[width=1.0\linewidth]{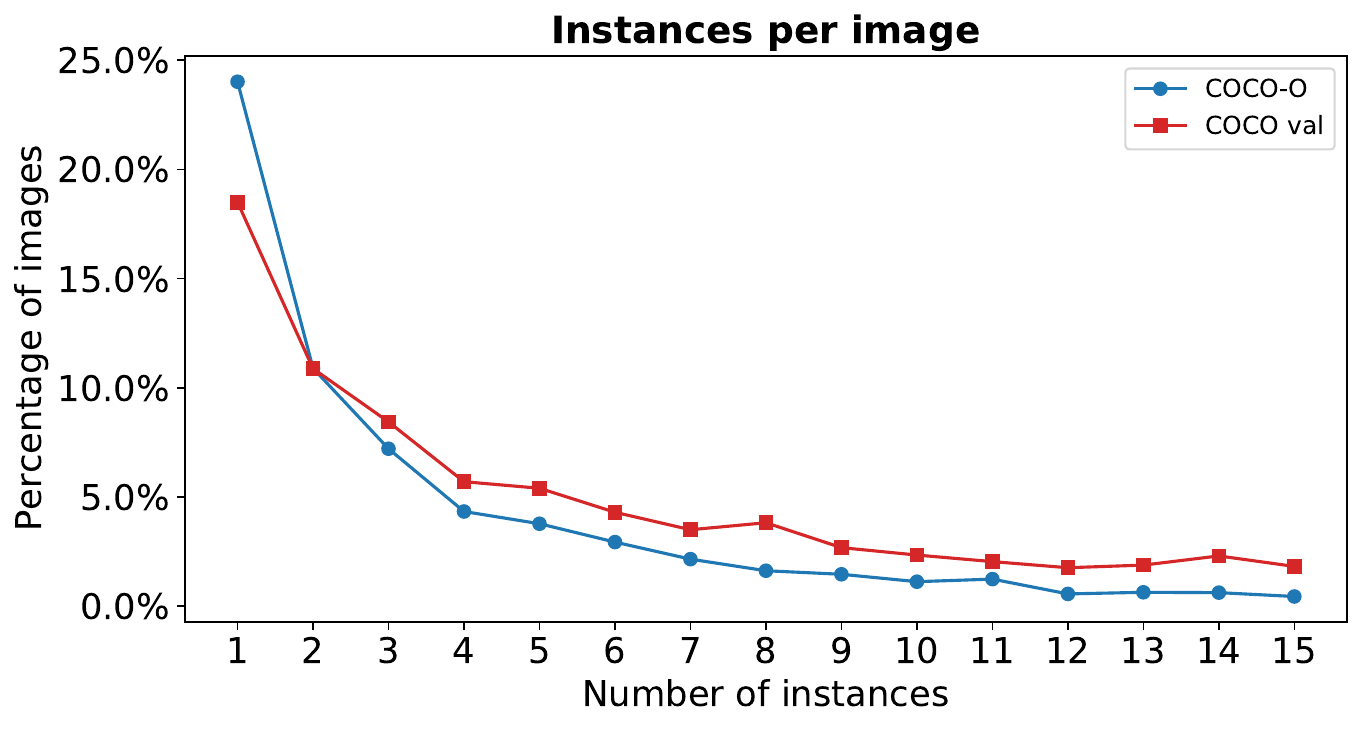}
  \caption{Number of annotated instances per image for COCO and COCO-O. }
  \label{fig:cocostat}
\end{figure}

\begin{figure*}[h!]
  \centering
  \includegraphics[width=0.95\linewidth]{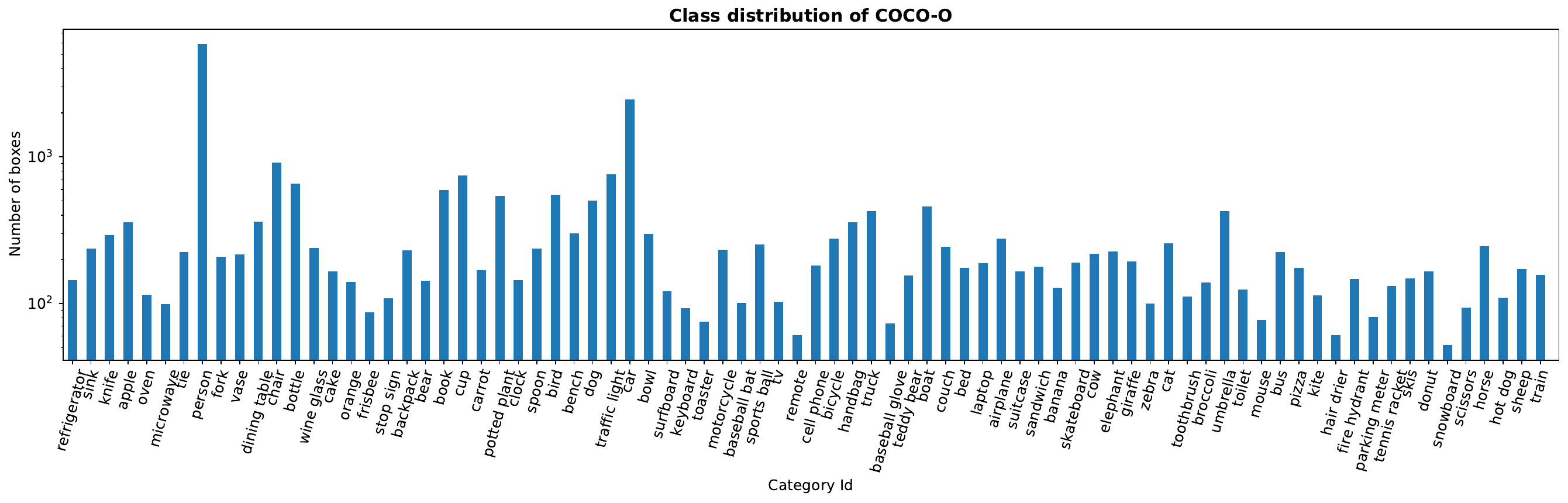}
  \vspace{-11pt}
  \caption{Class distribution of our COCO-O. }
  \label{fig:coco_class_dist}
  \vspace{-10pt}
\end{figure*}

We present the class distribution of COCO-O in Figure~\ref{fig:coco_class_dist}. The number of instances per image in COCO-O are shown in Figure~\ref{fig:cocostat}. COCO-O have around 5\% more iconic images compared with COCO validation set. To study the impact, we decrease the proportion by randomly deleting iconic images in COCO-O, and re-run the evaluation. The ablation in Table~\ref{tab:iconic} suggests more iconic images in COCO-O will not bring great change in result.

\begin{table}[h!]
    \small
    \centering
    \tablestyle{5pt}{1.05}
  \begin{tabular}{c|c|c|c|c}
  \toprule
  \multirow{2}{*}{Detectors} & \multicolumn{4}{c}{Percent of iconic images} \\
   & 24\% & 22\% & 20\% & 18\% \\
  \midrule
   Faster R-CNN & 16.4 & 16.3 & 16.5 & 16.4 \\
   RetinaNet & 16.6 & 16.5 & 16.4 & 16.5 \\
  \bottomrule
  \end{tabular}
  \caption{Impact of the more iconic images in COCO-O.}
  \label{tab:iconic}
\end{table}

\section{Detailed Results of Compared Detectors}
Table~\ref{tab:implementation} presents the detailed results of compared detectors in Figure~\ref{fig:main}. For compared SOTA detectors, we introduce their details in Table~\ref{tab:implementation_sota}. 

\section{Comparison of Results on COCO-C and COCO-O}
\begin{figure}[h!]
  \centering
  \includegraphics[width=1.0\linewidth]{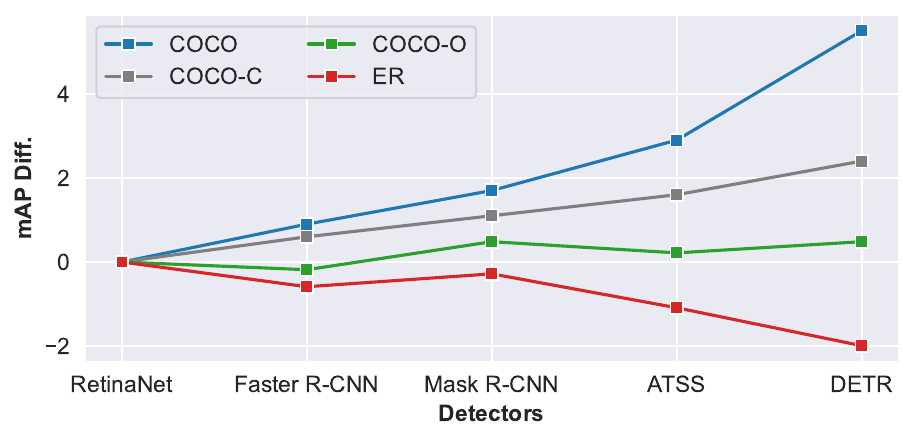}
  \vspace{-7mm}
  \caption{The results of COCO-C and COCO-O on different detectors. }
  \label{fig:coco-candcoco-o}
  \vspace{-5mm}
\end{figure}

Figure~\ref{fig:coco-candcoco-o} suggests COCO-C has a positive growth trend with clean performance, while such trend in COCO-O is more weak. It demonstrates COCO-O measures the robustness aspect out of standard COCO test set.

\section{Visualization of Detection Results}
\begin{figure*}[h!]
  \centering
  \includegraphics[width=1.0\linewidth]{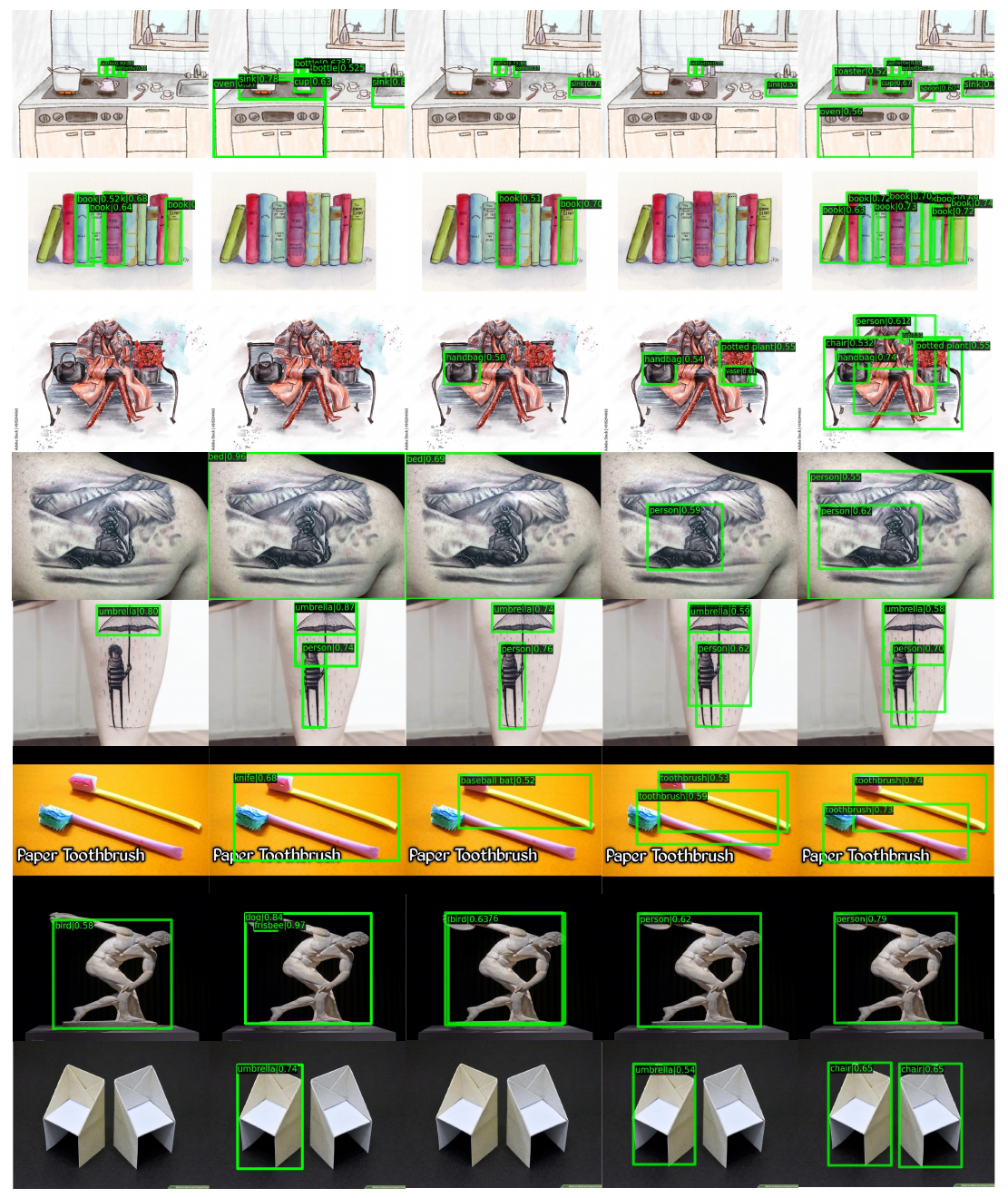}
  \caption{Visualization of detection results on sampled COCO-O images.}
  \label{fig:vis_sample2}
\end{figure*}

We adopt five detectors in Table~\ref{tab:implementation}: Faster R-CNN, DETR, YOLOX-S, PVTv2-B2, GLIP-T, and visualize their results on our COCO-O in Figure~\ref{fig:vis_sample2}. The score threshold is set as 0.5, which means that only bounding box with classification confidence over 0.5 will be visualized. Among the compared detectors, Faster R-CNN has failed to localize most objects and leads to a low recall rate. Detectors such as DETR, YOLOX-S recall some wrong bounding box with misclassified category or meaningless positions. GLIP-T outputs more accurate detection results. It can even recognize some abstract objects, such as the person with umbrella (the third line in figure).

\section{Visualization of Bad Cases}
\begin{figure*}[h!]
  \centering
  \includegraphics[width=1.0\linewidth]{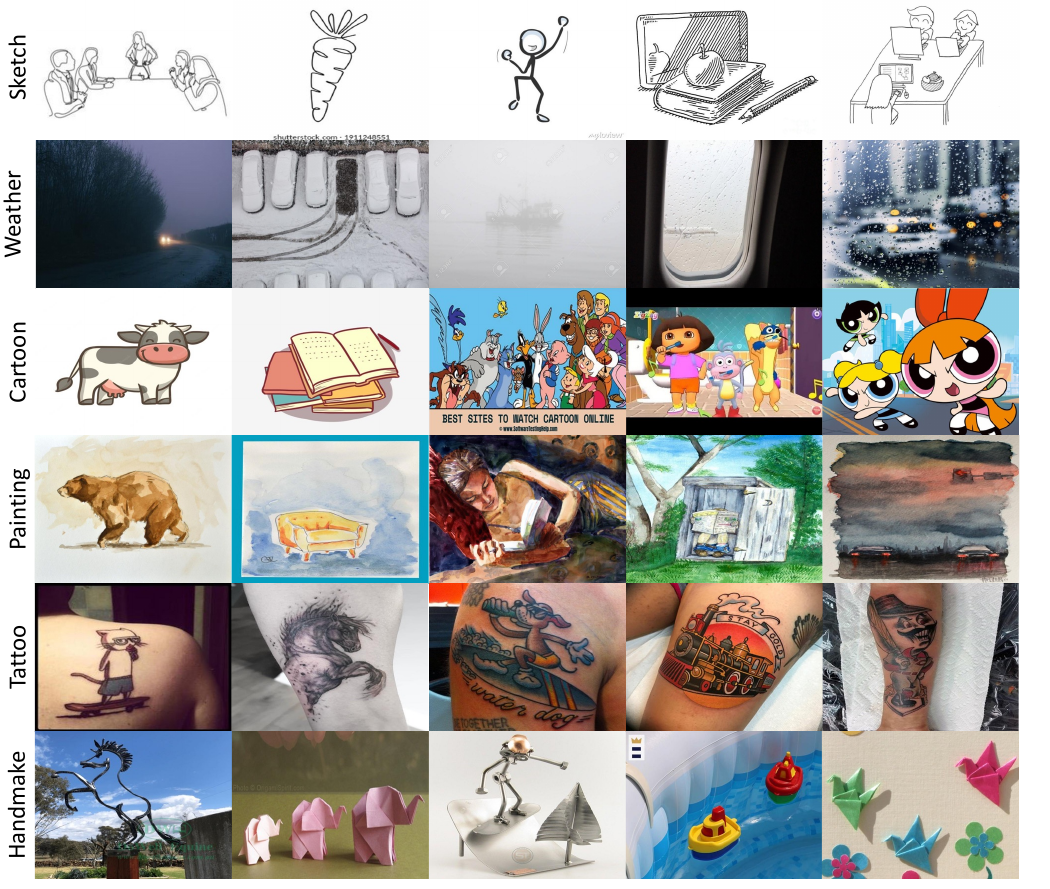}
  \caption{Hard examples in COCO-O which make detection failed.}
  \label{fig:badcase}
\end{figure*}

Among COCO-O images, we have found some bad cases which make most detectors fail. They have been shown Figure~\ref{fig:badcase}. Although the presented images contain common objects that can be easily recognized by humans, machines are hard to understand them. A vivid example is the anime characters in ``The Powerpuff Girls'' or ``Dora the Explorer'', which should be detected as person while machine looks unaware of such art form of expression. In addition, some traditional factors, \eg, perspective change, severe haze or blur, very poor lighting conditions, will also increase the difficulty of detection task.

\section{Usage of COCO-O}
\begin{itemize}
\item Example code for testing with MMDetection:
\end{itemize}
\begin{python}
$ git clone https://github.com/open-mmlab/
mmdetection.git
$ cd mmdetection
$ mkdir data
$ ln -s /path/to/coco-o/sketch data/coco
$ python tools/test.py ${CONFIG_FILE} \
${CHECKPOINT_FILE} --eval bbox
\end{python}
\begin{itemize}
\item Example code for testing with Detectron2: 
\end{itemize}
\begin{python}
$ git clone https://github.com/facebookresearch/
detectron2.git
$ cd detectron2
$ mkdir datasets
$ ln -s /path/to/coco-o/sketch datasets/coco
$ ./tools/lazyconfig_train_net.py \
--config-file ${CONFIG_FILE} \
train.init_checkpoint=${CHECKPOINT_FILE} \
--eval-only
\end{python}

\begin{table*}[h!]
\centering
    \small
    \centering
    \tablestyle{5pt}{1.05}
  \begin{tabular}{c|c|cc|cc|c|c|c|c|c}
  \toprule
  Detection & \multirow{2}{*}{BackBone} & COCO & COCO & COCO-O & COCO-O & Effective & Rank & Use & Additional & \multirow{2}{*}{FPS} \\
  Methods &  & mAP & rank & mAP & rank & Robustness & Diff. & Mask & Data & \\
  \midrule
  Mask R-CNN & ViT-B & 51.6 & 1 & 24.3 & 8 & \textcolor{green!50!black}{+1.06} & \textcolor{red}{-7} & \cmark & \xmark & 12.7 \\
  YOLOv7 & - & 51.2 & 2 & 29.0 & 3 & \textcolor{green!50!black}{+5.99} & \textcolor{red}{-1} & \xmark & \xmark & 161.0 \\
  FIBER-B & Swin-B & 49.5 & 3 & 33.7 & 1 & \textcolor{green!50!black}{+11.43} & \textcolor{green!50!black}{+2} & \xmark & \cmark & 3.1 \\
  Mask R-CNN & Swin-S & 48.5 & 4 & 25.5 & 6 & \textcolor{green!50!black}{+3.71} & \textcolor{red}{-2} & \cmark & \xmark & 10.1 \\
  Mask R-CNN & MViTV2-T & 48.3 & 5 & 25.7 & 5 & \textcolor{green!50!black}{+3.98} & 0 & \cmark & \xmark & 14.6 \\
  GLIP-T & Swin-T & 46.6 & 6 & 29.1 & 2 & \textcolor{green!50!black}{+8.11} & \textcolor{green!50!black}{+4} & \xmark & \cmark & 5.5 \\
  Mask R-CNN & Uniformer-S & 45.6 & 7 & 26.2 & 4 & \textcolor{green!50!black}{+5.65} & \textcolor{green!50!black}{+3} & \cmark & \xmark & 15.6 \\
  Mask R-CNN & PVTv2-B2 & 45.3 & 8 & 24.9 & 7 & \textcolor{green!50!black}{+4.52} & \textcolor{green!50!black}{+1} & \cmark & \xmark & 15.8 \\
  Deformable-DETR & ResNet-50 & 44.5 & 9 & 18.5 & 13 & \textcolor{red}{-1.49} & \textcolor{red}{-4} & \xmark & \xmark & 15.0 \\
  DyHead & ResNet-50 & 42.5 & 10 & 19.3 & 10 & \textcolor{green!50!black}{+0.16} & 0 & \xmark & \xmark & 13.2 \\
  HTC & ResNet-50 & 42.3 & 11 & 19.1 & 11 & \textcolor{green!50!black}{+0.08} & 0 & \cmark & \xmark & 5.8 \\
  DETR & ResNet-50 & 42.0 & 12 & 17.1 & 17 & \textcolor{red}{-1.82} & \textcolor{red}{-5} & \xmark & \xmark & 27.8 \\
  VarifocalNet & ResNet-50 & 41.6 & 13 & 18.8 & 12 & \textcolor{green!50!black}{+0.05} & \textcolor{green!50!black}{+1} & \xmark & \xmark & 19.4 \\
  YOLOX-S & - & 40.5 & 14 & 20.6 & 9 & \textcolor{green!50!black}{+2.48} & \textcolor{green!50!black}{+5} & \xmark & \xmark & 102.0 \\
  Cascade R-CNN & ResNet-50 & 40.3 & 14 & 18.2 & 14 & \textcolor{green!50!black}{+0.02} & 0 & \xmark & \xmark & 16.1 \\
  ATSS & ResNet-50 & 39.4 & 16 & 16.8 & 19 & \textcolor{red}{-0.91} & \textcolor{red}{-3} & \xmark & \xmark & 19.7 \\
  Mask R-CNN & ResNet-50 & 38.2 & 17 & 17.1 & 17 & \textcolor{red}{-0.11} & 0 & \cmark & \xmark & 16.1 \\
  Sparse R-CNN & ResNet-50 & 37.9 & 18 & 17.5 & 16 & \textcolor{green!50!black}{+0.41} & \textcolor{green!50!black}{+2} & \xmark & \xmark & 18.3 \\
  YOLOv5s & - & 37.4 & 19 & 17.8 & 15 & \textcolor{green!50!black}{+1.00} & \textcolor{green!50!black}{+4} & \xmark & \xmark & 156.3 \\
  Faster R-CNN & ResNet-50 & 37.4 & 19 & 16.4 & 22 & \textcolor{red}{-0.41} & \textcolor{red}{-3} & \xmark & \xmark & 21.4 \\
  FCOS & ResNet-50 & 36.6 & 21 & 16.7 & 20 & \textcolor{green!50!black}{+0.25} & \textcolor{green!50!black}{+1} & \xmark & \xmark & 22.7 \\
  RetinaNet & ResNet-50 & 36.5 & 22 & 16.6 & 21 & \textcolor{green!50!black}{+0.18} & \textcolor{green!50!black}{+1} & \xmark & \xmark & 19.0 \\
  YOLOS-S & - & 36.1 & 23 & 15.5 & 23 & \textcolor{red}{-0.80} & 0 & \xmark & \xmark & 5.7 \\
  YOLOv3 & DarkNet-53 & 33.7 & 24 & 14.8 & 24 & \textcolor{red}{-0.37} & 0 & \xmark & \xmark & 48.1 \\
  SSD & VGG16 & 29.5 & 25 & 13.6 & 25 & \textcolor{green!50!black}{+0.36} & 0 & \xmark & \xmark & 30.7 \\
  \bottomrule
  \end{tabular}
  \vspace{-10pt}
  \caption{Detailed results of the compared detectors in Figure~\ref{fig:main}.}
  \label{tab:implementation}
  \vspace{-15pt}
\end{table*}

\clearpage

\begin{table*}[h!]
\centering
    \small
    \centering
    \tablestyle{5pt}{1.05}
  \begin{tabular}{c|c|cc|cc|c|c|c|c|c}
  \toprule
  Detection & \multirow{2}{*}{BackBone} & COCO & COCO & COCO-O & COCO-O & Effective & Rank & Use & Additional & \multirow{2}{*}{FPS} \\
  Methods &  & mAP & rank & mAP & rank & Robustness & Diff. & Mask & Data & \\
  \midrule
EVA & - & 64.2 & 1 & 57.8 & 1 & \textcolor{green!50!black}{+28.86} & \textcolor{red}{0} & \cmark & \cmark & 0.8 \\
DETA & Swin-L & 62.9 & 2 & 48.5 & 2 & \textcolor{green!50!black}{+20.15} & \textcolor{red}{0} & \xmark & \cmark & 1.2 \\
GRiT & ViT-H & 60.4 & 3 & 42.9 & 4 & \textcolor{green!50!black}{+15.72} & \textcolor{red}{-1} & \cmark & \cmark & 0.7 \\
CBNetV2 & Swin-L & 59.1 & 4 & 39 & 7 & \textcolor{green!50!black}{+12.36} & \textcolor{red}{-3} & \cmark & \cmark & 2 \\
ViT-Adapter & BEiTv2-L & 58.8 & 5 & 34.25 & 13 & \textcolor{green!50!black}{+7.79} & \textcolor{red}{-8} & \cmark & \cmark & 5.6 \\
ViTDet & ViT-H & 58.7 & 6 & 34.3 & 12 & \textcolor{green!50!black}{+7.89} & \textcolor{red}{-6} & \cmark & \xmark & 4.8 \\
DINO & Swin-L & 58.5 & 7 & 42.1 & 5 & \textcolor{green!50!black}{+15.76} & \textcolor{green!50!black}{+2} & \xmark & \cmark & 1.8 \\
H-Deformable-DETR & Swin-L & 57.6 & 8 & 41.4 & 6 & \textcolor{green!50!black}{+15.45} & \textcolor{green!50!black}{+2} & \xmark & \cmark & 1.9 \\
YOLOv6-L6 & - & 57.2 & 9 & 32.5 & 18 & \textcolor{green!50!black}{+6.73} & \textcolor{red}{-9} & \xmark & \xmark & 26 \\
YOLOR-D6 & - & 57 & 10 & 33 & 17 & \textcolor{green!50!black}{+7.3} & \textcolor{red}{-7} & \xmark & \xmark & 34 \\
YOLOv7-E6E & - & 56.8 & 11 & 32 & 19 & \textcolor{green!50!black}{+6.42} & \textcolor{red}{-8} & \xmark & \xmark & 36 \\
SOLQ & Swin-L & 56.5 & 12 & 38.8 & 8 & \textcolor{green!50!black}{+13.41} & \textcolor{green!50!black}{+4} & \cmark & \cmark & 2.3 \\
DyHead & Swin-L & 56.2 & 13 & 35.3 & 11 & \textcolor{green!50!black}{+10} & \textcolor{green!50!black}{+2} & \xmark & \cmark & 6.3 \\
Cascade Mask R-CNN & InternImage-L & 56.1 & 14 & 37 & 10 & \textcolor{green!50!black}{+11.72} & \textcolor{green!50!black}{+4} & \cmark & \cmark & 9.1 \\
Cascade Mask R-CNN & MViTV2-H & 56.1 & 14 & 30.9 & 21 & \textcolor{green!50!black}{+5.62} & \textcolor{red}{-7} & \cmark & \cmark & 3.3 \\
CenterNet2 & R2-101-DCN & 56.1 & 14 & 29.5 & 25 & \textcolor{green!50!black}{+4.29} & \textcolor{red}{-11} & \xmark & \xmark & 11.8 \\
QueryInst & Swin-L & 55.5 & 17 & 33.2 & 16 & \textcolor{green!50!black}{+8.26} & \textcolor{green!50!black}{+1} & \cmark & \cmark & 3.3 \\
Cascade Mask R-CNN & ConvNeXt-XL & 55.2 & 18 & 37.5 & 9 & \textcolor{green!50!black}{+12.68} & \textcolor{green!50!black}{+9} & \cmark & \cmark & 8.6 \\
YOLOv4-P6 & - & 54.4 & 19 & 30.4 & 23 & \textcolor{green!50!black}{+5.89} & \textcolor{red}{-4} & \xmark & \xmark & 30 \\
YOLOv5l6 & - & 53.7 & 20 & 29.3 & 26 & \textcolor{green!50!black}{+5.17} & \textcolor{red}{-6} & \xmark & \xmark & 63 \\
Cascade Mask R-CNN & UniFormer-B & 53.4 & 21 & 31.8 & 20 & \textcolor{green!50!black}{+7.77} & \textcolor{green!50!black}{+1} & \cmark & \xmark & 12.4 \\
DiffusionDet & Swin-B & 52.2 & 22 & 34.2 & 14 & \textcolor{green!50!black}{+10.69} & \textcolor{green!50!black}{+8} & \xmark & \cmark & 10.8 \\
Det-AdvProp & EfficientNet-B5 & 52.2 & 22 & 30.8 & 22 & \textcolor{green!50!black}{+7.34} & \textcolor{red}{0} & \xmark & \xmark & 13.8 \\
Cascade Mask R-CNN & Swin-B & 51.9 & 24 & 25.5 & 35 & \textcolor{green!50!black}{+2.18} & \textcolor{red}{-11} & \cmark & \xmark & 11.6 \\
GLIP-L & Swin-L & 51.4 & 25 & 48 & 3 & \textcolor{green!50!black}{+24.89} & \textcolor{green!50!black}{+22} & \xmark & \cmark & 0.5 \\
EfficientDet-D5 & EfficientNet-B5 & 51.2 & 26 & 28.5 & 29 & \textcolor{green!50!black}{+5.44} & \textcolor{red}{-3} & \xmark & \xmark & 13.8 \\
YOLOX-X & - & 51.1 & 27 & 30.3 & 24 & \textcolor{green!50!black}{+7.26} & \textcolor{green!50!black}{+3} & \xmark & \xmark & 57.8 \\
Cascade Mask R-CNN & ResNeSt-200-DCN & 50.9 & 28 & 26.1 & 33 & \textcolor{green!50!black}{+3.18} & \textcolor{red}{-5} & \cmark & \xmark & 12.4 \\
UniverseNet & R2-101-DCN & 50.9 & 28 & 24.8 & 38 & \textcolor{green!50!black}{+1.86} & \textcolor{red}{-10} & \xmark & \xmark & 11.7 \\
HTC & RX-101-64x4d-DCN & 50.4 & 30 & 28.4 & 30 & \textcolor{green!50!black}{+5.67} & \textcolor{red}{0} & \cmark & \xmark & 2.1 \\
VFNet & RX-101-64x4d & 50.4 & 30 & 28 & 32 & \textcolor{green!50!black}{+5.27} & \textcolor{red}{-2} & \xmark & \xmark & 6.7 \\
GFLv2 & R2-101-DCN & 49.9 & 32 & 25.1 & 36 & \textcolor{green!50!black}{+2.6} & \textcolor{red}{-4} & \xmark & \xmark & 10.9 \\
FIBER-B & Swin-B & 49.5 & 33 & 33.7 & 15 & \textcolor{green!50!black}{+11.43} & \textcolor{green!50!black}{+18} & \xmark & \cmark & 3.1 \\
Mask R-CNN & MogaNet-L & 49.4 & 34 & 28.8 & 28 & \textcolor{green!50!black}{+6.57} & \textcolor{green!50!black}{+6} & \cmark & \xmark & 11.2 \\
RepPointsV2 & RX-101-64x4d-DCN & 49.3 & 35 & 24.9 & 37 & \textcolor{green!50!black}{+2.7} & \textcolor{red}{-2} & \xmark & \xmark & 5.8 \\
GCNet & RX-101-32x4d-DCN & 48 & 36 & 26 & 34 & \textcolor{green!50!black}{+4.38} & \textcolor{green!50!black}{+2} & \cmark & \xmark & 7.5 \\
TOOD & RX-101-64x4d & 47.6 & 37 & 23.1 & 40 & \textcolor{green!50!black}{+1.7} & \textcolor{red}{-3} & \xmark & \xmark & 6.1 \\
Mask R-CNN & PVTv2-B5 & 47.4 & 38 & 28.2 & 31 & \textcolor{green!50!black}{+6.85} & \textcolor{green!50!black}{+7} & \cmark & \xmark & 12.8 \\
Deformable DETR & ResNet-50 & 46.9 & 39 & 20.1 & 45 & \textcolor{red}{-1.01} & \textcolor{red}{-6} & \xmark & \xmark & 14.5 \\
GLIP-T & Swin-T & 46.6 & 40 & 29 & 27 & \textcolor{green!50!black}{+8.01} & \textcolor{green!50!black}{+13} & \xmark & \cmark & 5.5 \\
Cascade Mask R-CNN & RX-101-64x4d & 46.6 & 40 & 22.2 & 41 & \textcolor{green!50!black}{+1.23} & \textcolor{red}{-1} & \cmark & \xmark & 6.7 \\
HTC & HRNetV2-W40 & 46.4 & 42 & 24 & 39 & \textcolor{green!50!black}{+3.07} & \textcolor{green!50!black}{+3} & \cmark & \xmark & 4.3 \\
Mask R-CNN & RX-101-64x4d & 44.5 & 43 & 21.5 & 42 & \textcolor{green!50!black}{+1.48} & \textcolor{green!50!black}{+1} & \cmark & \xmark & 8 \\
Faster R-CNN & RX-101-64x4d & 43.1 & 44 & 20.2 & 44 & \textcolor{green!50!black}{+0.76} & \textcolor{red}{0} & \xmark & \xmark & 9.4 \\
FCOS & RX-101-64x4d & 42.6 & 45 & 21 & 43 & \textcolor{green!50!black}{+1.83} & \textcolor{green!50!black}{+2} & \xmark & \xmark & 9.7 \\
YOLOS-B & ViT-B & 42 & 46 & 20 & 46 & \textcolor{green!50!black}{+1.05} & \textcolor{red}{0} & \xmark & \xmark & 2.7 \\
DETR & ResNet-50 & 42 & 46 & 17.1 & 49 & \textcolor{red}{-1.82} & \textcolor{red}{-3} & \xmark & \xmark & 27 \\
RetinaNet & RX-101-64x4d & 41.6 & 48 & 19.9 & 47 & \textcolor{green!50!black}{+1.16} & \textcolor{green!50!black}{+1} & \xmark & \xmark & 8.7 \\
RepPointsV2 & MobileNetV2 & 36.8 & 49 & 16.4 & 51 & \textcolor{red}{-0.18} & \textcolor{red}{-2} & \xmark & \xmark & 32.1 \\
NanoDet & ShuffleNetV2 & 34.1 & 50 & 17.3 & 48 & \textcolor{green!50!black}{+1.92} & \textcolor{green!50!black}{+2} & \xmark & \xmark & 256 \\
YOLOv3 & DarkNet-53 & 33.7 & 51 & 14.8 & 52 & \textcolor{red}{-0.37} & \textcolor{red}{-1} & \xmark & \xmark & 48.1 \\
YOLACT & ResNet-101 & 30.4 & 52 & 16.8 & 50 & \textcolor{green!50!black}{+3.15} & \textcolor{green!50!black}{+2} & \cmark & \xmark & 33.5 \\
SSD & VGG16 & 29.5 & 53 & 13.6 & 53 & \textcolor{green!50!black}{+0.36} & \textcolor{red}{0} & \xmark & \xmark & 30.7 \\
  \bottomrule
  \end{tabular}
  \caption{Detailed results of the compared SOTA detectors in Figure~\ref{fig:sota}.}
  \label{tab:implementation_sota}
\end{table*}

\clearpage

\clearpage

\begin{figure*}[h!]
  \centering
  \includegraphics[width=1.0\linewidth]{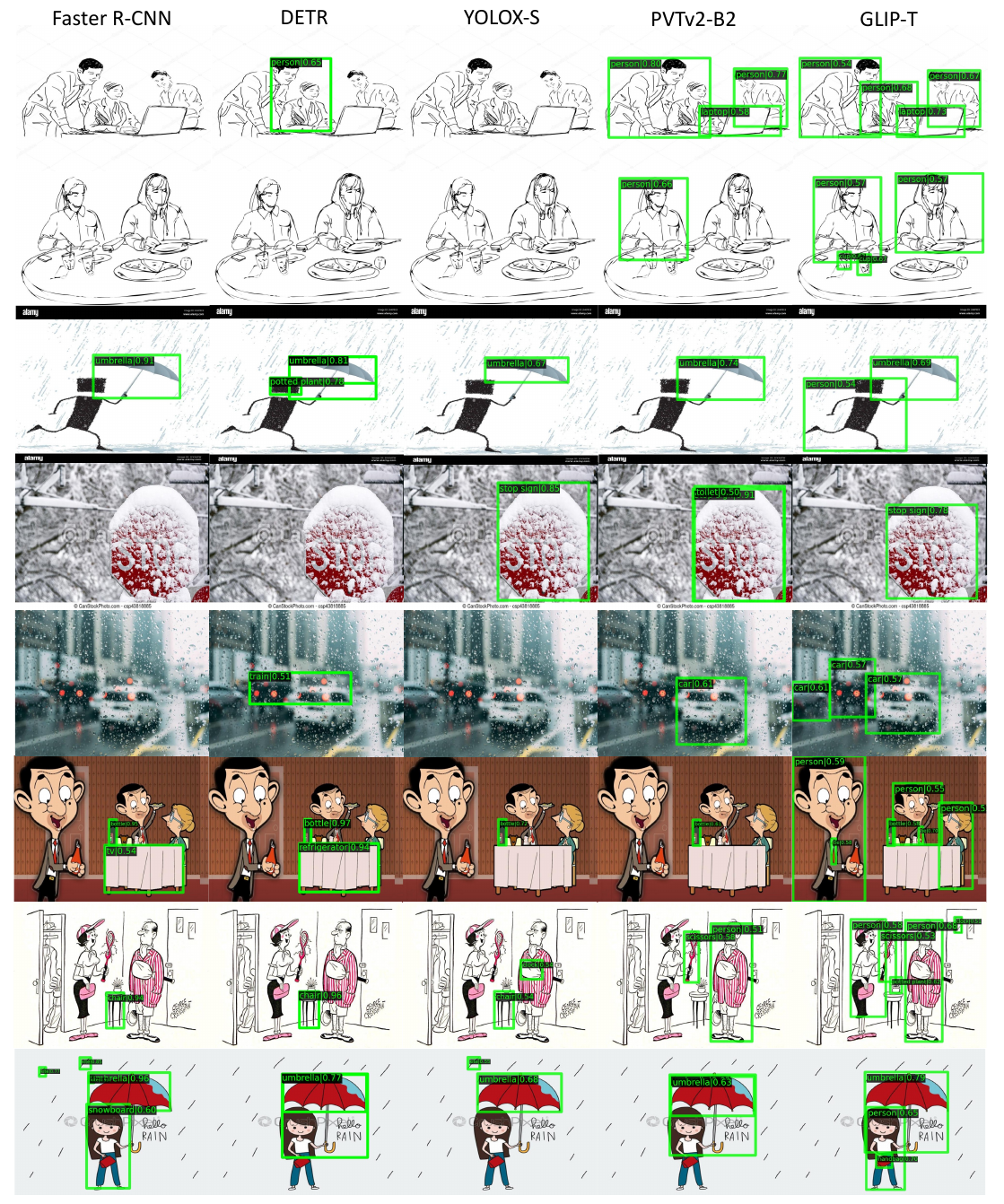}
  \caption*{}
  \label{fig:vis_sample}
\end{figure*}

\clearpage

\clearpage

\section{Queries for Collecting COCO-O}
We show the query keywords used for collect sketch images in COCO-O. Similarly, for other test domains, we create queries by just replacing with the corresponding domain keywords. 
\begin{python}
sketch person
sketch bicycle
sketch car
sketch motorcycle
sketch airplane
sketch bus
sketch train
sketch truck
sketch boat
sketch traffic light
sketch fire hydrant
sketch stop sign
sketch parking meter
sketch bench
sketch bird
sketch cat
sketch dog
sketch horse
sketch sheep
sketch cow
sketch elephant
sketch bear
sketch zebra
sketch giraffe
sketch backpack
sketch umbrella
sketch handbag
sketch tie
sketch suitcase
sketch frisbee
sketch skis
sketch snowboard
sketch sports ball
sketch kite
sketch baseball bat
sketch baseball glove
sketch skateboard
sketch surfboard
sketch tennis racket
sketch bottle
sketch wine glass
sketch cup
sketch fork
sketch knife
sketch spoon
sketch bowl
sketch banana
sketch apple
sketch sandwich
sketch orange
sketch broccoli
sketch carrot
sketch hot dog
sketch pizza
sketch donut
sketch cake
sketch chair
sketch couch
sketch potted plant
sketch bed
sketch dining table
sketch toilet
sketch tv
sketch laptop
sketch mouse
sketch remote
sketch keyboard
sketch cell phone
sketch microwave
sketch oven
sketch toaster
sketch sink
sketch refrigerator
sketch book
sketch clock
sketch vase
sketch scissors
sketch teddy bear
sketch hair drier
sketch toothbrush
sketch person chair
sketch person car
sketch person handbag
sketch person dining table
sketch chair dining table
sketch person cup
sketch person backpack
sketch person bottle
sketch cup dining table
sketch person cell phone
sketch person tennis racket
sketch person bench
sketch person truck
sketch person sports ball
sketch car truck
sketch bowl dining table
sketch person umbrella
sketch person bus
sketch person surfboard
sketch person tie
sketch person bowl
sketch cup chair
sketch bottle cup
sketch person motorcycle
sketch person skateboard
sketch person bicycle
sketch bottle dining table
sketch person skis
sketch knife dining table
sketch cup bowl
sketch car traffic light
sketch person traffic light
sketch fork dining table
sketch car bus
sketch person knife
sketch bottle chair
sketch person book
sketch person baseball glove
sketch bottle bowl
sketch person remote
sketch person baseball bat
sketch person pizza
sketch person couch
sketch chair couch
sketch chair book
sketch person horse
sketch spoon bowl
sketch spoon dining table
sketch cup knife
sketch person tv
sketch chair tv
sketch backpack handbag
sketch fork knife
sketch person kite
sketch cup spoon
sketch sports ball tennis racket
sketch person wine glass
sketch person laptop
sketch bowl chair
sketch person clock
sketch chair potted plant
sketch person train
sketch pizza dining table
sketch person dog
sketch person suitcase
sketch person boat
sketch person potted plant
sketch person spoon
sketch bottle sink
sketch person cake
sketch person frisbee
sketch person fork
sketch cup fork
sketch cake dining table
sketch mouse keyboard
sketch toilet sink
sketch tv book
sketch tv keyboard
sketch chair laptop
sketch wine glass dining table
sketch laptop keyboard
sketch couch remote
sketch couch book
sketch laptop mouse
sketch car motorcycle
sketch car handbag
sketch baseball bat baseball glove
sketch chair vase
sketch bottle knife
sketch potted plant dining table
sketch couch tv
sketch knife chair
sketch knife bowl
sketch person bed
sketch cup pizza
sketch wine glass cup
sketch tv mouse
sketch fork bowl
sketch bottle wine glass
sketch couch dining table
sketch tv laptop
sketch bicycle car
sketch potted plant vase
sketch dining table vase
sketch sandwich dining table
sketch umbrella handbag
sketch laptop book
sketch person chair dining table
sketch person cup dining table
sketch person car truck
sketch cup chair dining table
sketch person cup chair
sketch person bottle dining table
sketch person car bus
sketch person backpack handbag
sketch person sports ball tennis racket
sketch cup bowl dining table
sketch person car traffic light
sketch person bottle cup
sketch person bottle chair
sketch person knife dining table
sketch fork knife dining table
sketch person bowl dining table
sketch bottle chair dining table
sketch bottle cup dining table
sketch bowl chair dining table
sketch cup knife dining table
sketch cup fork dining table
sketch person cup bowl
sketch person car handbag
sketch person baseball bat baseball glove
sketch cup spoon dining table
sketch person bottle bowl
sketch person fork dining table
sketch spoon bowl dining table
sketch person pizza dining table
sketch tv mouse keyboard
sketch person car motorcycle
sketch person umbrella handbag
sketch person cup knife
sketch knife chair dining table
sketch bottle cup bowl
sketch bottle bowl dining table
sketch cup fork knife
sketch person spoon dining table
sketch person cup spoon
sketch cup spoon bowl
sketch person fork knife
sketch chair potted plant dining table
sketch person spoon bowl
sketch person bicycle car
sketch laptop mouse keyboard
sketch person cup fork
sketch person car backpack
sketch person wine glass dining table
sketch person cake dining table
sketch person wine glass cup
sketch bottle cup chair
sketch fork chair dining table
sketch chair couch dining table
sketch person couch remote
sketch person knife chair
sketch person bowl chair
sketch chair couch tv
sketch fork bowl dining table
sketch person tennis racket chair
sketch person handbag chair
sketch person bottle wine glass
sketch chair dining table vase
sketch person chair book
sketch tv laptop keyboard
sketch cup pizza dining table
sketch person bottle knife
sketch bottle knife dining table
sketch car truck traffic light
sketch person sports ball baseball glove
sketch pizza chair dining table
sketch person cup pizza
sketch cup bowl chair
sketch chair dining table book
sketch knife bowl dining table
sketch person bicycle backpack
sketch chair tv book
sketch person wine glass chair
sketch person fork chair
sketch tv laptop mouse
sketch person handbag cell phone
sketch person bus truck
sketch person umbrella chair
sketch person pizza chair
sketch bottle fork dining table
sketch wine glass cup dining table
sketch person handbag suitcase
sketch chair couch book
sketch knife spoon dining table
sketch person chair laptop
sketch fork pizza dining table
sketch bottle cup knife
sketch person bicycle handbag
sketch car bus truck
sketch fork spoon dining table
sketch person bench handbag
sketch bottle spoon bowl
sketch person chair potted plant
sketch wine glass chair dining table
sketch cup fork chair
sketch person fork pizza
sketch person chair cell phone
sketch person bottle spoon
sketch cup knife chair
sketch chair potted plant vase
sketch person cup cake
sketch chair couch potted plant
sketch person bus handbag
sketch person car bench
sketch person sports ball baseball bat
sketch person sports ball chair
\end{python}

\end{document}